\documentclass[10pt,twocolumn,letterpaper]{article}

\usepackage{cvpr}      

\usepackage[usenames,dvipsnames]{xcolor}
\usepackage{times}
\usepackage{float}
\usepackage{booktabs}
\usepackage{cellspace}
\usepackage{epsfig}
\usepackage{graphicx}
\usepackage{amsmath}
\usepackage{amssymb}
\usepackage[nice]{nicefrac}
\usepackage{mathtools}
\usepackage{multirow}
\usepackage{enumitem}
\usepackage{color}
\usepackage{dsfont}
\usepackage{algorithm}
\usepackage{algpseudocode}
\usepackage{makecell}
\usepackage{lipsum}
\usepackage{amsmath}
\usepackage{wrapfig}
\usepackage{bm}
\usepackage{pifont} 
\usepackage{multirow}
\usepackage{colortbl}
\definecolor{Gray}{gray}{0.9}
\usepackage[pagebackref,breaklinks,colorlinks]{hyperref}
\usepackage[capitalize]{cleveref}
\crefname{section}{Sec.}{Secs.}
\Crefname{section}{Section}{Sections}
\Crefname{table}{Table}{Tables}
\crefname{table}{Tab.}{Tabs.}

\def\cvprPaperID{6974} 
\def\confName{CVPR}
\def\confYear{2023}

\newcommand{\keypoint}[1]{\vspace{0.01cm}\noindent\textbf{#1}\quad}
\newcommand{\cut}[1]{}

\definecolor{deepGreen}{RGB}{0,153,0}
\definecolor{myBlue}{RGB}{108,142,191}
\definecolor{newcolor}{RGB}{133,99,99} 

\algnewcommand\algorithmicforeach{\textbf{for each}}
\algdef{S}[FOR]{ForEach}[1]{\algorithmicforeach\ #1\ \algorithmicdo}

\begin{document}
\title{\vspace{-0.5cm} CLIP for All Things Zero-Shot Sketch-Based Image Retrieval, \\Fine-Grained or Not}

\vspace{-0.5cm}
\author{
Aneeshan Sain\textsuperscript{1,2}  \hspace{.2cm} 
Ayan Kumar Bhunia\textsuperscript{1} \hspace{.3cm}  
Pinaki Nath Chowdhury\textsuperscript{1,2} \hspace{.3cm}
Subhadeep Koley\textsuperscript{1,2} \hspace{.3cm}\\
Tao Xiang\textsuperscript{1,2}\hspace{.4cm}  
Yi-Zhe Song\textsuperscript{1,2} \\
\textsuperscript{1}SketchX, CVSSP, University of Surrey, United Kingdom.  \\
\textsuperscript{2}iFlyTek-Surrey Joint Research Centre on Artificial Intelligence.\\
{\tt\small \{a.sain, a.bhunia, p.chowdhury, s.koley, t.xiang, y.song\}@surrey.ac.uk} 
\vspace{-0.2cm}
}

\maketitle
\begin{abstract}
{
In this paper, we leverage CLIP for zero-shot sketch based image retrieval (ZS-SBIR). We are largely inspired by recent advances on foundation models and the unparalleled generalisation ability they seem to offer, but for the first time tailor it to benefit the sketch community. We put forward novel designs on how best to achieve this synergy, for both the category setting and the fine-grained setting (``all''). At the very core of our solution is a prompt learning setup. First we show just via factoring in sketch-specific prompts, we already have a category-level ZS-SBIR system that overshoots all prior arts, by a large margin ($24.8\%$) -- a great testimony on studying the CLIP and ZS-SBIR synergy. Moving onto the fine-grained setup is however trickier, and requires a deeper dive into this synergy. For that, we come up with two specific designs to tackle the fine-grained matching nature of the problem: (i) an additional regularisation loss to ensure the relative separation between sketches and photos is \textit{uniform} across categories, which is not the case for the gold standard standalone triplet loss, and (ii) a clever patch shuffling technique to help establishing instance-level structural correspondences between sketch-photo pairs. With these designs, we again observe significant performance gains in the region of $26.9\%$ over previous state-of-the-art. The take-home message, if any, is the proposed CLIP and prompt learning paradigm carries great promise in tackling other sketch-related tasks (not limited to ZS-SBIR) where data scarcity remains a great challenge. Project page: \url{https://aneeshan95.github.io/Sketch_LVM/}
}
\end{abstract}

\vspace{-0.6 cm}
\section{Introduction}
\vspace{-0.2cm}
\label{sec:intro}

\cut{Song's points:
1. CLIP for Everything SBIR: Fine-grained, Or not
2. 1b -- Prompt for SBIR , Prompt for FG-SBIR
}

Late research on sketch-based image retrieval (SBIR) \cite{sain2020cross,sain2021stylemeup,sain2023exploiting} had fixated on the zero-shot setup, i.e., zero-shot SBIR (ZS-SBIR) \cite{dey2019doodle,yelamarthi2018zero,dutta2019semantically}. This shift had become inevitable because of data-scarcity problem plaguing the sketch community \cite{bhunia2021more,bhunia2023sketch2saliency,koley2023picture} -- there are just not enough sketches to train a  general-purpose SBIR model. It follows that the key behind a successful ZS-SBIR model lies with how best it conducts semantic transfer cross object categories \textit{and} between sketch-photo modalities. Despite great strides made elsewhere on the general zero-shot literature \cite{actionclip2021,clip-continual2022,zhou2022learning}  however, semantic transfer \cite{DenseCLIP} for ZS-SBIR had remained rather rudimentary, mostly using standard word embeddings {directly \cite{dey2019doodle,dutta2019semantically,zhang2020zero} or indirectly \cite{tian2021relationship,liu2019semantic,wang2022prototype}}.

In this paper, we fast track ZS-SBIR research to be aligned with the status quo of the zero-shot literature, and for the first time, propose a synergy between foundation models like CLIP\cite{clip} and the cross-modal problem of ZS-SBIR. And to demonstrate the effectiveness of this synergy, we not only tackle the conventional category-level ZS-SBIR, but a new and more challenging fine-grained instance-level \cite{pang2020solving} ZS-SBIR as well. 

Our motivation behind this synergy of CLIP and ZS-SBIR is no different to the many latest research adapting CLIP to vision-language pre-training \cite{gao2021clip}, image and action recognition \cite{DenseCLIP,actionclip2021} and especially on zero-shot tasks \cite{maple2022,clip-continual2022,luo2021clip4clip,zhou2022learning} -- CLIP exhibits a highly enriched semantic latent space, and already encapsulates knowledge across a myriad of cross-modal data. As for ZS-SBIR, CLIP is therefore almost a perfect match, as (i) it already provides a rich semantic space to conduct category transfer, and (ii) it has an unparalleled understanding on multi-modal data, which SBIR dictates.

\begin{figure}[!t]
\centering
\includegraphics[width=\linewidth]{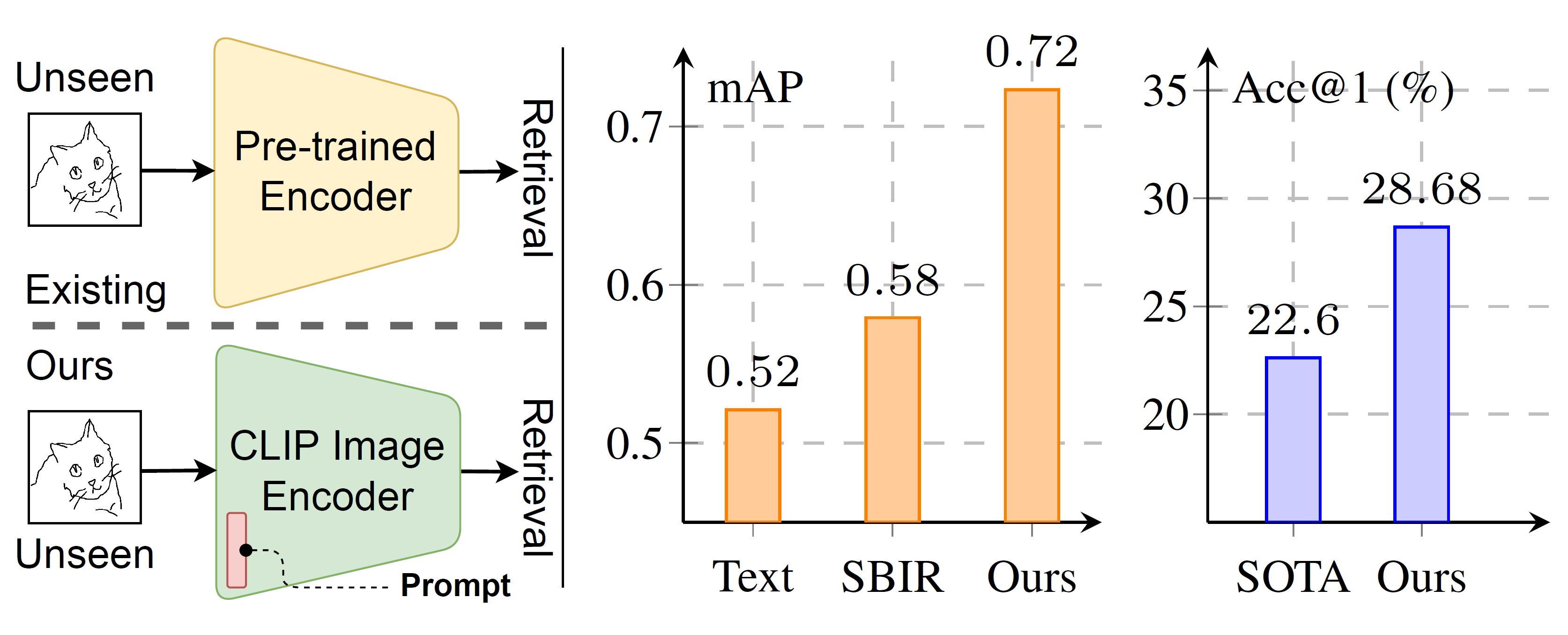}
\vspace{-0.80cm}
\caption{{Against existing (left) ZS-SBIR methods, we adapt CLIP model for ZS-SBIR (middle), and extend to a more practical yet challenging setup of FG-ZS-SBIR (right), via a novel prompt-based design. Our model surpasses prior arts by a high margin.}}
\label{fig:teaser}
\vspace{-0.6cm}
\end{figure}

At the very heart of our answer to this synergy is that of prompt learning \cite{jia2022visual}, {that involves learning a set of continuous vectors injected into CLIP's encoder. This enables CLIP to adapt to downstream tasks while preserving its generalisability -- a theme that we follow in our CLIP-adaption to ZS-SBIR (\cref{fig:teaser}).} {More specifically, we first design two sets of \textit{visual} prompts, one for each modality (sketch, photo). They are both injected into the initial layer insider the transformer of CLIP for training. While keeping the rest of CLIP frozen, these two prompts are trained over the gold standard triplet loss paradigm \cite{yu2016sketch}, on extracted sketch-photo features. } Motivated by the efficacy of training batch normalisation for image recognition \cite{frankle2021batchNorm}, we additionally fine-tune a small subset of trainable parameters of every Layer Normalisation (LN) layer for additional performance gain. {Furthermore, to enhance cross-category semantic transfer, we also resort to CLIP's text encoder further cultivating its zero-shot potential. In particular, in addition to the said \textit{visual} prompts onto the CLIP's image encoder, we use handcrafted \textit{text} prompts via templates like `\textit{photo of a [category]}' to its text encoder, during training. }

The new fine-grained setting \cite{pang2020solving,chowdhury2023what} is however more tricky. Unlike the previous category-level setup, it poses two additional challenges (i) relative feature-distances between sketch-photo pairs across categories are non-uniform, which is reflected in the varying triplet-loss margin \cite{yu2016sketch} across categories at training \cite{bhunia2022adaptive}, and (ii) apart from semantic consistency, fine-grained ZS-SBIR requires instance-level matching to be conducted \cite{pang2020solving}, which dictates additional constraints such as structural correspondences. 

It follows that for the first challenge, we propose a new regularisation term that aims at making the relative sketch-photo feature-distances uniform across categories, such that a single (global) margin parameter works across all of them. Specifically, taking the distribution of relative distances for all sketch-positive-negative hard-triplets \cite{yu2016sketch} in a category, we aim to minimise the KL-divergence \cite{matthews2016sparse} between every pair of distributions, which trains the model towards making such relative sketch-photo distances uniform across categories. For the latter, we propose a clever patch shuffling technique, where equally divided patches of a sketch and its \textit{corresponding} photo (\textbf{n}$\times$\textbf{n}) are first{ shuffled following a random permutation order of patches. We then advocate that a shuffled sketch should be closer to a shuffled photo having the same permutation order, but far from that of a different permutation.}
Training this permutation-invariance imparts a broad notion of structural correspondences, thus helping in fine-grained understanding. 

{Summing up: (i) We for the first time adapt CLIP for ZS-SBIR, (ii) We propose a novel prompt learning setup to facilitate the synergy between the two, (iii) We address both the conventional ZS-SBIR setting, and a new and more challenging fine-grained ZS-SBIR problem, (iv) We introduce a regularisation term and a clever patch shuffling technique to address the fine-grained challenges. With our CLIP-adapted model surpassing all prior arts by a large margin (\cref{fig:teaser}), we hope to have shed some light to the sketch community on benefits such a synergy between foundation models and sketch-related tasks can bring. }

\vspace{-0.25cm}
\section{Related Work}
\label{sec:related}
\vspace{-0.2cm}
                                                                        
\keypoint{Category-level SBIR:} Given a query-sketch, SBIR aims at fetching category-specific photos from a gallery of multi-category photos. Recent deep-frameworks aim to learn a joint sketch-photo manifold via a feature extractor ~\cite{collomosse2019livesketch, dey2019doodle, xu2018sketchmate, xu2018sketchmate} over a triplet-ranking objective~\cite{yu2016sketch}. Towards practicality of \textit{unseen} test-time classes, Zero-Shot SBIR (ZS-SBIR) was explored for cross-category generalisation~\cite{dey2019doodle,yelamarthi2018zero}, and enhanced via test-time training~\cite{sain2022sketch3t}. Others explored binary hash-codes~\cite{liu2017deep, zhang2018generative} for computational ease. \textit{Sketch} however specializing in modelling \textit{fine-grained} details, geared research towards \textit{Fine-Grained} SBIR.

\keypoint{Fine-grained SBIR:} FG-SBIR aims at retrieving \textit{one} instance from a gallery of \textit{same}-category images based on a query-sketch. Introduced as a deep triplet-ranking based \emph{siamese network}~\cite{yu2016sketch} for learning a joint sketch-photo manifold, FG-SBIR was improvised via attention-based modules with a higher order retrieval loss~\cite{song2017deep}, textual tags~\cite{song2017fine,chowdhury2023scenetrilogy}, hybrid cross-domain generation~\cite{pang2017cross}, hierarchical co-attention~\cite{sain2020cross} and reinforcement learning~\cite{bhunia2020sketch}. Furthermore, sketch-traits like style-diversity~\cite{sain2021stylemeup}, data-scarcity~\cite{bhunia2021more} and redundancy of sketch-strokes~\cite{bhunia2022sketching} were addressed in favor of retrieval. Towards generalising to novel classes, while \cite{pang2019generalising} modelled a universal manifold of prototypical visual sketch traits embedding sketch and photo, \cite{bhunia2022adaptive} adapted to new classes via some supporting sketch-photo pairs. In this paper, we aim to address the problem of zero-shot cross-category FG-SBIR, leveraging the zero-shot potential of a foundation model like CLIP \cite{clip}. 

\keypoint{Zero-Shot SBIR:} This aims at generalising knowledge learned from \textit{seen} training classes to \textit{unseen} testing categories. Yelamarthi \etal \cite{yelamarthi2018zero} introduced ZS-SBIR, to reduce sketch-photo domain gap by approximating photo features from sketches via image-to-image translation. While, \cite{dutta2019semantically} aligned sketch, photo and semantic representations via adversarial training, \cite{dey2019doodle} minimised sketch-photo domain gap over a gradient reversal layer. Improvising further, others used graph convolution networks \cite{zhang2020zero}, or distilled contrastive relationships \cite{tian2021relationship} in a student from ImageNet-pretrained teacher, coupled skecth/photo encoders with shared conv layers and independent batchnorm-layer \cite{wang2021transferable}, a shared ViT \cite{tian2022tvt} to minimise domain gap, or employed prototype-based selective knowledge distillation \cite{wang2022prototype} on learned correlation matrix, and very recently introduced a test-time training paradigm \cite{sain2022sketch3t} via reconstruction on test-set sketches, adapting to the test set distribution. {Semantic transfer for ZS-SBIR however was mostly limited to using word embeddings directly \cite{dutta2019semantically,zhang2020zero,wang2021transferable} or indirectly  \cite{tian2021relationship,liu2019semantic,wang2022prototype}.} Furthermore, FG-ZS-SBIR being non-trivial remains to be explored. In this paper, we thus take to adapting CLIP to exploit its high generalisability for semantic transfer and exploring its zero-shot potential for FG-ZS-SBIR.

\keypoint{CLIP in Vision Tasks:} Contrastive Language-Image Pre-training (CLIP) \cite{clip} trains on cross-modal data, benefiting both from rich semantic textual information \cite{DenseCLIP} and large scale availability of images ($\sim 400$M image-text pairs) for training. Unlike traditional representations on discretized labels, CLIP represents images and text in the same embedding space \cite{DenseCLIP}, thus enabling generalizability in downstream tasks with no labels (zero-shot) \cite{zhou2022conditional} or a few annotations (few-shot) \cite{zhou2022learning} by generating classification weights from text encodings. Efficiently adapting CLIP for downstream tasks to exploit it's zero shot potential has been investigated using Prompt engineering from NLP literature \cite{liu2022NLPprompt}. Such common extensions include retrieval \cite{baldrati2022effective}, image generation \cite{text2light2022}, continual learning \cite{clip-continual2022}, object detection,  few-shot recognition \cite{gao2021clip}, semantic segmentation \cite{DenseCLIP}, etc. Others leveraged CLIP's image/text encoders with StyleGAN \cite{patashnik2021styleclip} to enable intuitive text-based semantic image manipulation \cite{abdal2022clip2stylegan}. In our work we adapt CLIP for zero-shot FG-SBIR in a cross-category setting.

\keypoint{Prompt Learning for Vision Tasks:}\label{sec:prompt} Originating from NLP domain, prompting \cite{brown2020language} imparts context to the model regarding the task at hand, thus utilising the knowledge base of large-scale pretrained text models like GPT and BERT to benefit downstream tasks.  This involves constructing a task specific template (e.g., \texttt{`The movie was [MASK]'}), and label words (e.g. \texttt{`good/bad'}) to fill it up. Domain-expertise being necessary for hand-crafted prompt-engineering, urged the need for prompt tuning in recent works \cite{jia2022visual}, which entails modelling the prompt as task-specific learnable continuous vectors that are directly optimised via gradient descent during fine-tuning. Learning context vectors for prompting has also taken root in the vision community \cite{bahng2022visual}, with greater emphasis on prompting large scale vision language models \cite{zhou2022conditional} or visual feature extractors (e.g. ViT \cite{jia2022visual}). Unlike previous attempts at prompt learning \cite{zhou2022learning,jia2022visual}, in context of FG-SBIR, we focus on learning a single prompt for sketch and photo branches which when used with CLIP, would generalise on to unseen novel classes by leveraging the zero-shot potential of CLIP for multi-category FG-ZS-SBIR.

\vspace{-0.25cm}
\section{Preliminaries} \label{sec:prelim}
\vspace{-0.25cm}
\keypoint{Overview of CLIP:} 
Contrastive Language-Image Pre-training or CLIP  \cite{clip}, widely popular for open-set visual understanding tasks, consists of two separate encoders, one for image and another one for text.  The image encoder ($\mathbf{V}$) uses either ResNet-50 \cite{HeResNet} or a Vision Transformer (ViT) \cite{dosovitskiy2021image}, where an input image ($p \in \mathbb{R}^{H \times W \times 3}$) is divided into $m$ fixed-size patches and embedded as $E_{0}$ = $\{\mathbf{e}_{0}^{j} \}_{j=1}^m; \mathbf{e}_{0}^{j} \in \mathbb{R}^{d_p} $. Similar to BERT's [$\texttt{CLASS}$] token, a learnable class token $c_0^t \in \mathbb{R}^d_p$ is appended, and the resultant matrix $[E_{0}, c_0^v] \in \mathbb{R}^{(m+1)\times d}$ is passed through transformer layers, followed by a feature projection layer on class-token feature to obtain the final \emph{visual feature} $f_p = \mathbf{V}(p) \in \mathbb{R}^d$ in joint vision-language embedding space. 
Similarly, using a vocab-size of $49,152$, the text encoder $\mathbf{T}$ first converts a sentence with $n$ words (including punctuation) to word embeddings as $W_{0}$ = $\{\mathbf{w}_{0}^{j}\}_{j=1}^n \; ; \mathbf{w}_{0}^{j} \in \mathbb{R}^{d_t}$ and appends a learnable class token $c_0^t \in \mathbb{R}^{d_t}$ to form the input feature matrix $[W_{0}, c_0^t] \in \mathbb{R}^{(n+1) \times d_t}$ representing knowledge of the sentence ($\mathcal{S}$), which is passed via a transformer to extract textual feature $f_{t} = \mathbf{T}(\mathcal{S})$. The model is trained via a contrastive loss \cite{clip}  maximising cosine similarity for matched text-photo pairs while minimising it for all other unmatched pairs. During downstream tasks like classification \cite{clip},  textual prompts like ‘\texttt{a photo of a [category]}’ (from a list of $K$ categories) are fed to $\mathbf{T}$ to obtain category-specific text features  ($f_t$) and calculate the prediction probability for input photo feature ($f_p$) as:
\vspace{-0.2cm}
\begin{equation}
    \label{eq:primary}
    \mathcal{P}(y|p) = \frac{\mathrm{exp}(\texttt{sim}(f_p, f_t^y)/\tau)}{\sum_{i=1}^{K} \mathrm{exp}(\texttt{sim}(f_p, f_t^i)/\tau)}
\vspace{-0.15cm}
\end{equation}

\vspace{0.05cm}
\keypoint{Prompt Learning:} 
Following NLP literature \cite{brown2020language}, prompt learning has been adopted on foundation models like CLIP \cite{zhou2022conditional} or large-pretrained Vision Transformers \cite{jia2022visual} to benefit downstream tasks. Unlike previous fine-tuning based methods \cite{pang2020solving} that entirely updated  weights of pre-trained models on downstream task datasets, prompt learning keeps weights of foundation models frozen to avoid deploying separate copy of large models for individual tasks besides preserving pre-learned generalizable  knowledge.
For a visual transformer backbone, concatenated patch-features and class tokens ($[E_0, c_0^v] \in \mathbb{R}^{(m+1)\times d_p}$), appended with a set of $K$ learnable prompt vectors $\mathbf{v}^p$ = $\{v_i\}_{i=1}^{K}; v_i \in \mathbb{R}^{d_p}$ for photo $p$, as $[E_0, c_0^v, \mathbf{v}^p ]$ $\in \mathbb{R}^{(m+1+K)\times d_p}$, and passed via the transformer to obtain $\widehat{f_p}$ = $\mathbf{V}(p, \mathbf{v}^p)$.
Keeping the entire model fixed, $\mathbf{v}^p$ is fine-tuned on the task-specific dataset, adapting the foundation model to the task at hand. Variations of prompt learning include, shallow vs deep prompt \cite{jia2022visual}, based on the layers at which prompts are inserted. Here, we use simple shallow prompt that is inserted only at the first layer along-with patch embeddings, which empirically remains optimal and easy to reproduce.

\vspace{-0.2cm}
\section{CLIP for Zero-Shot \textit{Category-level} SBIR} \label{sec:ZS-CLIP}
\vspace{-0.15cm}
\keypoint{Baseline Categorical SBIR:}
Given a query-sketch ($s$) from any category, categorical SBIR~\cite{sain2022sketch3t} aims to retrieve a photo of the \textit{same} category, from a gallery ($\mathcal{G}$) holding photos from multiple ($N_c$) categories $\mathcal{G}$ = $\{ p^j_i \}_{i=1}^{M_i}|_{j=1}^{N_c}$, where $i^\text{th}$ class has $M_i$ number of photos. Formally, an embedding (separate for sketch and photo) function $\mathcal{F}(\cdot) : \mathbb{R}^{H\times W\times 3} \rightarrow \mathbb{R}^d$, usually represented by an ImageNet\cite{krizhevsky2012imagenet}-pretrained VGG-16 \cite{Simonyan15}, is trained to extract a $d$-dimensional feature from an input image (sketch $s$ /photo $p$) $\mathcal{I}$ as $f_\mathcal{I} = \mathcal{F}_{\mathcal{I}}(\mathcal{I}) \in \mathbb{R}^d$, over a triplet loss \cite{yu2016sketch} using the feature triplet of a query-sketch ($f_s$) and a photo ($f_p$) belonging to the same category, and another photo from a different category ($f_n$). Minimising the triplet loss \cut{over entire sketch-dataset} ($\mathcal{L}_\text{Tri}$) signifies bringing sketches and photos of the same category closer while distancing other categories' photos. With $\mu$ as margin and distance function $d(a,b)$ = $({1-a\cdot b})/({||a||-||b||})$, triplet loss is given as,

\vspace{-0.2cm}
\begin{equation}\label{eq: cat-triplet}
    \mathcal{L}_{\text{Tri}} = \max \{0, \mu + d(f_s, f_p) - d(f_s, f_n) \}
\end{equation}
\vspace{-0.5cm}

Unlike normal SBIR, evaluating on categories \textit{seen} during training $\mathcal{C}^\text{S}$ = $\{c^\text{S}_i\}^{N_\text{S}}_{i=1}$, ZS-SBIR \cite{dey2019doodle} evaluates on novel ones $\mathcal{C}^\text{U} $ = $\{c^\text{U}_i\}^{N_\text{U}}_{i=1}$, \textit{unseen} during training, \textit{i.e}, $\mathcal{C}^\text{S} \cap \mathcal{C}^\text{U} = \emptyset$.

\vspace{0.1cm}
\keypoint{Naively Adapting CLIP for ZS-SBIR:}
Although the SBIR \textit{baseline} can be naively extended to a zero-shot setting for ZS-SBIR \cite{dey2019doodle}, it performs unsatisfactorily lacking sufficient zero-shot transfer \cite{clip-continual2022,luo2021clip4clip} of semantic knowledge. Consequently, a very naive extension upon CLIP would be to replace ImageNet\cite{russakovsky2015imagenet}-pretrained VGG-16 by the CLIP's visual encoder, which already holds semantic-rich information, thus directly harnessing its inherent zero-shot potential for ZS-SBIR. Following traditional fine-tuning methods \cite{pang2020solving}, if we start naively training the CLIP's image encoder on SBIR dataset using triplet loss, the performance collapses due to \textit{catastrophic forgetting} \cite{sain2022sketch3t} of learnt CLIP knowledge. Alternatively, focusing on parameter-efficient paradigm, one may train via an additional MLP layer at the end \cite{gao2021clip}, while keeping the remaining network frozen. As this essentially transfers the encoded feature from CLIP's embedding space to a subsequent space via the MLP layer, it does not guarantee CLIP's generalisation potential to remain preserved \cite{jia2022visual}, thus defeating our sole purpose of adapting CLIP for ZS-SBIR. Avoiding such loopholes, we therefore opt for a prompt learning approach \cite{zhou2022conditional} that not only provides stable optimisation \cite{kim2022how} but also preserves the desired generalisation (open-vocab) \cite{zhou2022learning} of CLIP.

\vspace{0.1cm}
\keypoint{Prompt Learning for ZS-SBIR:} To adapt CLIP for category-level SBIR,
we learn two sets of sketch/photo visual prompts as $\mathbf{v}^{s}, \mathbf{v}^{p} \in \mathbb{R}^{K \times d_p}$, each of which is injected into respective sketch $\mathcal{F}_s$ and photo $\mathcal{F}_p$ encoders both initialised from CLIP's image encoder, respectively. Finally, we get the prompt\footnote{Please see \S Prompt Learning in \cref{sec:prelim} and Supplementary for details.} guided sketch feature $f_{s}$ = $\mathcal{F}_{s}(s, \mathbf{v}^{s}) \in \mathbb{R}^{d}$ and photo feature $f_{p}$ = $\mathcal{F}_{p}(p, \mathbf{v}^{p}) \in \mathbb{R}^{d}$, respectively. In essence, the sketch and photo specific prompts \emph{induce} CLIP \cite{clip} to learn the downstream sketch and photo distribution respectively. Knowledge learned by CLIP is distilled into a prompt's weights via backpropagation keeping CLIP visual encoder's weights ($\theta$) frozen. While freezing $\theta$ is motivated by training stability  \cite{zhou2022learning, zhou2022conditional}, we take a step further and ask, can we improve CLIP by fine-tuning $\theta$ yet enjoy training stability? Accordingly, instead of fine-tuning $\theta$ entirely, we tune a small subset -- the trainable parameters of every layer normalisation (LN) layers across $\theta$. Our design is motivated by prior observation \cite{frankle2021batchNorm} on unusual efficacy of training batch normalisation for image recognition, keeping rest of parameters frozen. Therefore, besides the prompt parameters $\{\mathbf{v}^{s}, \mathbf{v}^{p}\}$, we update the parameters of sketch/photo branch specific layer-norm layers's parameters $\{l_\theta^s, l_\theta^p\}$ via standard triplet loss as in~\cref{eq: cat-triplet}. The trainable parameter set is $\{\mathbf{v}^{s}, \mathbf{v}^{p}, l_\theta^s, l_\theta^p\}$.

\vspace{+0.05cm}
\keypoint{Classification Loss using CLIP's Text Encoder:}
Besides using CLIP's image encoder for zero-shot utility, we further exploit the high generalisation ability provided by natural language \cite{devlin2018bert, clip} through CLIP's text encoder. In particular, along with the triplet loss, we impose a classification loss on the sketch/photo joint-embedding space. For this, instead of usual auxiliary $N_s$-class FC-layer based classification head \cite{bhunia2022adaptive, Dutta_2019_CVPR, dey2019doodle}, we take help of CLIP's text encoder to compute the classification objective, which is already enriched with semantic-visual association. Following \cite{gu2022KD}, we construct a set of handcrafted prompt templates like `\texttt{a photo of a [category]}' to obtain a list classification weight vectors $\{t_j\}_{j=1}^{N_s}$  using CLIP's text encoder where the `\texttt{[category]}' token is filled with a specific class name from a list of $N_s$ seen classes. 
The classification loss for $\mathcal{I}= \{s, p\}$ is given by:

\vspace{-0.6cm}
\begin{equation}
    \begin{aligned}
        \mathcal{L}^{\mathcal{I}}_{\text{cls}} &= \frac{1}{N}\sum_{i=1}^N - \log \mathcal{P} (y_i | I_i) \quad \text{where,}\\[-1pt]
        \mathcal{P}(y_i|I_i) &= \frac{\text{exp}(\texttt{sim}( \mathcal{F}_{\mathcal{I}}(\mathcal{I}_i),t_y)/\tau)}{\sum^{N_s}_{j=1} \text{exp}(\texttt{sim}(\mathcal{F}_{\mathcal{I}_i}(\mathcal{I}),t_j)/\tau)}
    \end{aligned}
    \label{eq: ZS-text}
\end{equation}
\vspace{-0.30cm}

\noindent Summing up our CLIP adapted ZS-SBIR paradigm is trained using a weighted ($\lambda_1$) combination of losses as : $\mathcal{L}^\text{ZS-SBIR}_\text{Trn} = \mathcal{L}_\text{Tri} + \lambda_1 (\mathcal{L}^p_\text{cls} + \mathcal{L}^s_\text{cls})$. 

While off-the-shelf CLIP itself has zero-shot image retrieval potential \cite{maple2022,clip-continual2022} where someone can feed the category level query as `\texttt{a photo of a [query]}', it raises a question -- how much is a category-level query-sketch beneficial over text-keyword based query? Attending to sketch's specialty in modelling fine-grained \cite{sain2020cross,yu2016sketch,bhunia2022sketching} details hence, we go beyond category-level ZS-SBIR \cite{dey2019doodle,dutta2019semantically} to a more practical and long-standing research problem of cross-category fine-grained ZS-SBIR \cite{pang2020solving}.

\vspace{-0.2cm}
\section{CLIP for Zero-Shot \textit{Fine-grained} SBIR}
\vspace{-0.2cm}
\keypoint{Background on FG-SBIR:} Compared to category-level SBIR \cite{sain2022sketch3t}, fine-grained SBIR \cite{yu2016sketch} aims at instance-level sketch-photo matching at intra-category level. Most of existing FG-SBIR works remain restricted to single-category setup, where they train and evaluate on the same category, like the standard FG-SBIR dataset (e.g., QMUL-ShoeV2 \cite{yu2016sketch}), that comprises $k$ instance-level sketch/photo pairs as $\{s_i, p_i\}_{i=1}^{k}$. A baseline FG-SBIR framework \cite{yu2016sketch} involves training a backbone network, shared between sketch and photo branches using a triplet-loss based objective \cite{yu2016sketch} where the matched sketch-photo pairs respectively form the anchor ($s_i$) and positive ($p_i$) samples, whereas a random photo $(p_{\neq i})$ is considered as the negative. 

A few works have extended it to multi-category FG-SBIR \cite{bhunia2022adaptive} setup which aims to train a single model with instance-level matching from multiple ($N_c$) categories (e.g., Sketchy dataset \cite{sangkloy2016sketchy}). The dataset consists of sketch/photo pairs from multiple categories $\{s_i^j, p_i^j\}_{i=1}^{k_j} |_{j=1}^{N_c}$ with every $j^\text{th}$ class having $k_j$ sketch-photo pairs. On top of baseline for single-category FG-SBIR \cite{sain2020cross}, it involves two additional design considerations \textit{(i)} Hard-triplets for triplet loss based training where the negative photo $(p_{\neq i}^{j})$ is from the same $j^\text{th}$ class of sketch-anchor ($s_{i}^{j}$) and positive-photo ($p_{i}^{j}$), but of different instances ($\mathcal{L}^\text{hard}_\text{Tri}$), (ii) an auxiliary $N_c$-class classification head on the sketch/photo joint embedding space to learn the class discriminative knowledge. 

Moving on, cross-category zero-shot FG-SBIR \cite{pang2019generalising} is analogous to category-level ZS-SBIR, in that the training and testing categories are disjoint ($\mathcal{C}^\text{S} \cap \mathcal{C}^\text{U} = \emptyset$), but the former needs to fetch instance-level photos from unseen categories instead of merely retrieving at category level like the latter. We therefore aim to answer the question: how can we extend our CLIP-based ZS-SBIR to FG-ZS-SBIR?

\begin{figure}[!htbp]
    \vspace{-0.2cm}
    \centering
    \includegraphics[width=\linewidth]{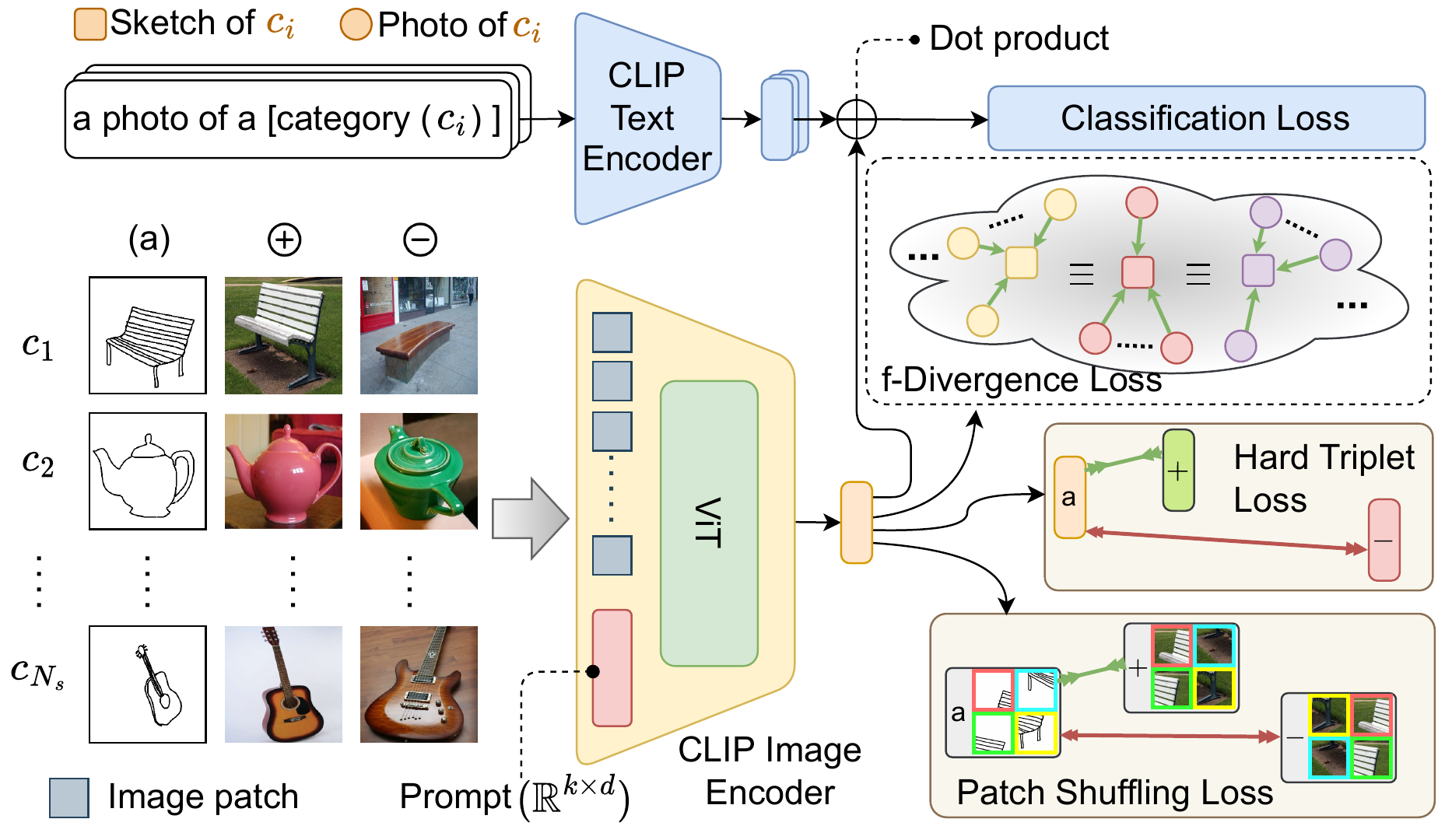}
    \vspace{-0.7cm}
    \caption{Cross-category FG-ZS-SBIR. A common (photo-sketch) learnable visual prompt shared across categories is trained using CLIP's image encoder over three losses as shown. CLIP's text-encoder based classification loss is used during training.}
    \label{fig:framework}
    \vspace{-0.4cm}
\end{figure}

\keypoint{Extending CLIP-based ZS-SBIR to FG-ZS-SBIR:} 
To recap (\cref{fig:framework}), the key components of CLIP-based ZS-SBIR are: (i) CLIP image-encoder as backbone with separate sketch/photo branches with individual prompts $\{\mathbf{v}^s, \mathbf{v}^p\}$, (ii) category-level triplets (iii) CLIP text-encoder \cite{clip} based classification loss, and (iv) fine tuning layer-norms for sketch/photo branches. Keeping rest of the design same, the necessary modification for intra-category instance-level matching (FG-ZS-SBIR) is to replace the \textit{category-level} triplets by \textit{hard}-triplets -- ($s_{i}^{j}, p_{i}^{j}, p_{\neq i}^{j}$) all from the \textit{same} category but \textit{different} negative instances. Furthermore, we empirically found that a common prompt \cite{jia2022visual} and a shared backbone \cite{yu2016sketch} between sketch/photo branches works better for fine-grained matching. The only \textit{trainable} parameter set is thus a common prompt $\mathbf{v}$ and layer-norm parameters $l_\theta$.

However, there are two major bottlenecks: Firstly, due to instance level matching across categories, the category-specific margin-parameter of triplet loss ($\mu$) varies significantly \cite{bhunia2022adaptive}, showing that a single global margin-value alone is sub-optimal for training a FG-ZS-SBIR model. 
Secondly, due to the diverse shape morphology \cite{sain2021stylemeup} amongst varying categories, it becomes extremely challenging to recognise fine-grained associations for unseen classes whose shape is \textit{unknown}. We therefore need a training signal to explicitly learn the structural correspondences in a sketch-photo pair.

\keypoint{Stabilising Margin ($\mu$) across Categories:}
Recently a work \cite{bhunia2022adaptive} on multi-category FG-SBIR has empirically shown optimal margin ($\mu$) value to vary across different categories.  Looking closely at triplet loss, $\mathcal{L} = \max(0, \mu + d(s, p^{+}) - d(s, p^{-}))$ \cite{yu2015sketch}, it essentially computes the difference between positive ($(s, p^{+})$) and negative distances ($(s, p^{-})$). We term this difference as the \emph{relative distance} $\delta(s, p^{+}, p^{-}) = d(s, p^{+}) - d(s, p^{-})$. Using a \textit{constant} $\mu$ means that on average, the relative distance is same for any triplet $(s, p^{+}, p^{-})$ \cite{song2017deep} in a category. Contrarily, a varying $\mu$ across different categories signifies that the average relative distance across categories is not uniform \cite{bhunia2022adaptive}. Therefore, naively training with a single $\mu$ value across all seen categories would be sub-optimal, and affect the \textit{cross-category generalisation} of triplet loss \cite{pang2020solving} which importantly works on this relative distance. While \cite{bhunia2022adaptive} tackles this issue by meta-learning \cite{tim2020metaSurvey} the margin value using few-shot sketch/photo pairs, ours is entirely a zero-shot setup \cite{dey2019doodle}, rendering such adaptation infeasible.  We thus impose a regulariser that aims to make this relative distance \textit{uniform} across categories such that the same triplet loss, with single (global) margin parameter $\mu$ works for all categories. To achieve this, we first compute the distribution of relative distances \cite{long2008, garcia2012} for all triplets $(s, p^{+}, p^{-})$ in category $c$ as $\mathcal{D}_{c} = \mathrm{softmax}\{\delta(s_i, p^{+}_i, p^{-})\}_{i=1}^{N_s}$, where $c^{th}$ category has $N_s$ sketch-photo pairs. Next, towards making the relative distance uniform across categories, we minimise the KL-divergence \cite{matthews2016sparse} between a distribution of relative distances between every category-pair (aka. f-divergence \cite{garcia2012}) as:
\vspace{-0.3cm}
\begin{equation}\label{eq: f-divergence}
    \mathcal{L}_{\delta} = \frac{1}{N_s(N_s-1)} \sum_{i=1}^{N_s} \sum_{j=1}^{N_s} \texttt{KL} (\mathcal{D}_{i}, \mathcal{D}_{j})
\vspace{-0.2cm}
\end{equation}
In practice, we compute $\mathcal{L}_{\delta}$  using sketch/photo samples from every category appearing in a batch. Importantly the spread, or relative entropy \cite{shannon1948} or information radius \cite{sibson1969} of distribution $\delta$  should be similar, thus stabilising training with a single margin value for multi-category FG-ZS-SBIR.
 
\begin{figure}[!htbp]
    \vspace{-0.1cm}
    \centering
    \includegraphics[width=\linewidth]{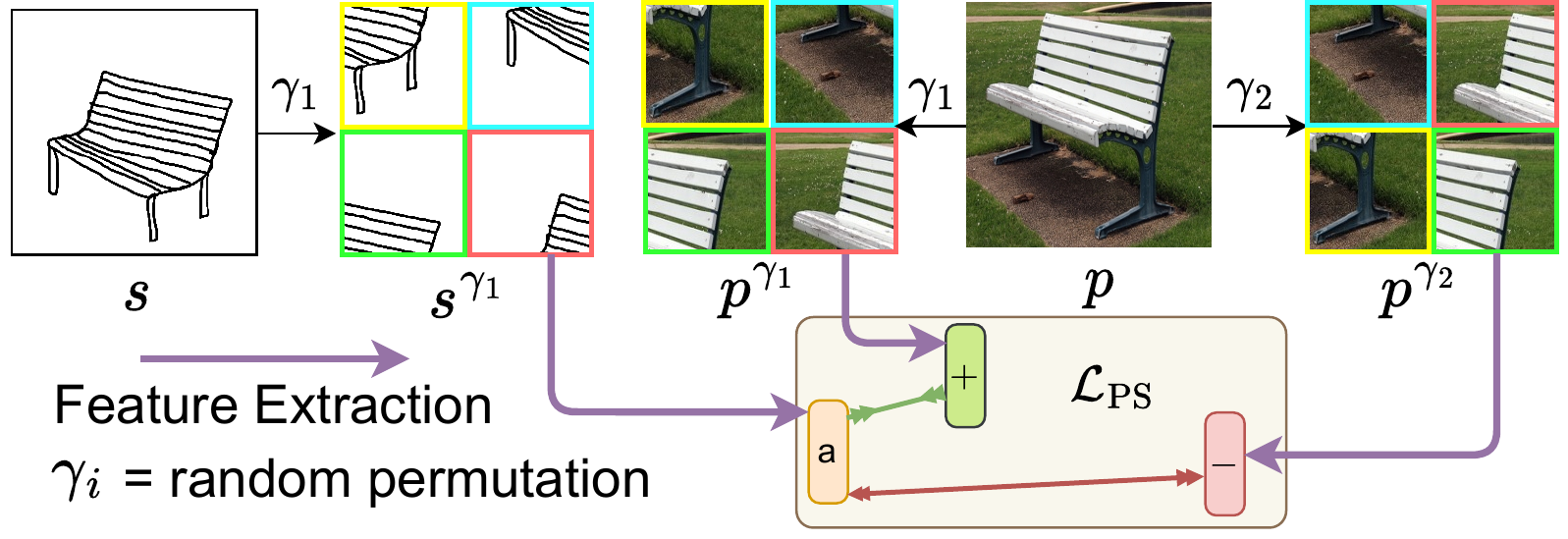}
    \vspace{-0.7cm}
    \caption{Patch-shuffling for fine-grained transfer}.
    \label{fig:patch_shuffle}
    \vspace{-0.8cm}
\end{figure}

\keypoint{Patch-shuffling for Zero-Shot Fine-grained Transfer:} 
Category-level SBIR is subtly different from FG-SBIR \cite{yu2016sketch} in that the former focuses only on semantic similarity between sketch-photo pairs, unlike FG-SBIR that takes a step further to focus on fine-grained shape matching \cite{bhunia2022adaptive} between sketches and photos. Highly diverse shape morphology \cite{sain2021stylemeup} across new categories implies unconstrained domain gap for multi-category FG-SBIR, thus increasing its difficulty. Discovering fine-grained correspondence becomes even harder as shape itself becomes unknown for unseen categories \cite{bhunia2022adaptive}.

For better fine-grained shape-matching transfer to novel classes, we design a simple data-augmentation trick through patch-shuffling to create augmented triplets \cite{song2017deep}. In particular,  we permute $\mathbf{n}\times \mathbf{n}$ patches (numbered) of sketch ($s$) and photo ($p$) using $\psi(\cdot)$ as $s^{\gamma}$ = $\psi(s, \gamma)$ and $p^\gamma$ = $\psi(p, \gamma)$, where $\gamma$ denotes a random permutation of the array $[1,2,...\mathbf{n}^2]$ describing the mapping of image patches to $s^{\gamma}$ or $p^{\gamma}$ (\cref{{fig:patch_shuffle}}). Given a sketch-photo pair of any category ($s, p$), training should decrease feature-distance of the sketch-permutation ($s^{\gamma_1}$) from the same permutation ($\gamma_1$) of its paired photo ($p^{\gamma_1}$), while increasing it from a different permutation ($p^{\gamma_2}$). Accordingly, we devise a triplet \cite{yu2016sketch} training objective as :

\vspace{-0.6cm}
\begin{equation}\label{eq: shuffling}
    \mathcal{L}_{\text{PS}} = \max \{0, \mu_{ps} + d(f_{s^{\gamma_1}}, f_{p^{\gamma_1}}) - d(f_{s^{\gamma_1}}, f_{p^{\gamma_2}}) \}
\vspace{-0.15cm}
\end{equation}

In contrast to auxiliary patch-order prediction, we found that our triplet objective between similar and dissimilar permuted instances provides better fine-grained shape transfer, besides being much cheaper during trainsing compared to complex Sinkorn operation \cite{adams2011sinkhorn} as used by Pang \etal \cite{pang2020solving}.

With $\lambda_{2,3,4}$ as hyperparameters, our overall training objective for CLIP adapted FG-ZS-SBIR paradigm is given as, $\mathcal{L}^\text{FG-ZS-SBIR}_\text{Trn} = \mathcal{L}_\text{Tri}^\text{hard} + \lambda_2 (\mathcal{L}_\text{cls}^{s} + \mathcal{L}_\text{cls}^{p}) + \lambda_3 \mathcal{L}_{\delta} +\lambda_4 \mathcal{L}_\text{PS}$.

\vspace{-0.15cm}
\section{Experiments}
\label{sec:expt}
\vspace{-0.2cm}

\keypoint{Datasets:} 
We use three popular datasets for evaluation on ZS-SBIR.
(i) \textbf{Sketchy (extended)} \cite{liu2017deep} --  Sketchy \cite{sangkloy2016sketchy} contains 75,471 sketches over 125 categories having 100 images each \cite{yelamarthi2018zero}. 
We use its extended version \cite{liu2017deep} having extra 60,502 images from ImageNet \cite{russakovsky2015imagenet}.
Following \cite{yelamarthi2018zero} we split it as 104 classes for training and 21 for testing for zero-shot setup.
(ii) \textbf{TUBerlin \cite{eitz2012humans}} -- contains 250 categories, with 80 free-hand sketches in each, extended to a total of 204,489 images by \cite{zhang2016sketchnet}. Following \cite{dey2019doodle} We split it as 30 classes for testing and 220 for training. (iii) \textbf{QuickDraw Extended} \cite{quickdraw}-- The \textit{full}-version houses over 50 million sketches across 345 categories. Augmenting them with images, a subset with 110 categories having 330,000 sketches and 204,000 photos was introduced for ZS-SBIR \cite{dey2019doodle}, which we use, following their split of 80 classes for training and 30 for testing. Requiring fine-grained sketch-photo association \cite{song2017deep} for evaluating cross-category FG-ZS-SBIR, we resort to Sketchy \cite{sangkloy2016sketchy} with \textit{fine-grained sketch-photo association}, using the same zero-shot categorical split of 104 training and 21 testing classes \cite{yelamarthi2018zero}.

\keypoint{Implementation Details:}
{We implemented our method in PyTorch on a 11GB Nvidia RTX 2080-Ti GPU. For sketch/photo encoder, we use CLIP \cite{clip} with ViT backbone using ViT-B/32 weights. For both paradigms of ZS-SBIR and FG-ZS-SBIR the input image size is set as $224 \times 224$ with margin parameter $\mu$=$0.3$, and prompts are trained using Adam optimiser with learning rate $1e-5$ for $60$ epochs, and batch size $64$, while keeping CLIP model fixed except its LayerNorm layers. We use two prompts (sketch and photo) from ZS-SBIR and one common prompt for FG-ZS-SBIR, each having a dimension of $(3 \times 768)$. Our prompts are injected in the first layer of transformer. For FG-ZS-SBIR $\mathbf{n}$=2 patches are used for patch shuffling-objective. Values of $\lambda_{1,2,3,4}$ are set to $0.5$, $0.5$, $0.1$ and $1$, empirically.
}
 
\keypoint{Evaluation Metric:}
Following recent ZS-SBIR literature \cite{dey2019doodle,hwang2020variational,wang2022prototype} we perform ZS-SBIR evaluation considering the top 200 retrieved samples, reporting mAP score (mAP@all) and precision (P@200) for ZS-SBIR. Keeping consistency with recent ZS-SBIR works however, we report P@100 and map@200 specifically for TUBerlin~\cite{zhang2016sketchnet} and Sketchy-ext \cite{liu2017deep} respectively. For cross-category FG-ZS-SBIR, accuracy is measured taking only a single category at a time \cite{pang2020solving}, as Acc.@q \cite{yu2016sketch} for Sketchy \cite{sangkloy2016sketchy}, which reflects percentage of sketches having true matched photo in the top-q list. We use Top-1 and Top-5 lists \cite{bhunia2022adaptive}.

{
\begin{table*}[!htbp]
    \centering
    \renewcommand{\arraystretch}{0.85}
    \scriptsize
    \caption{Quantitave comparison of our method against existing frameworks and baselines on ZS-SBIR and cross-category FG-ZS-SBIR.}
    \vspace{-0.3cm}
    \setlength{\aboverulesep}{0pt}
    \setlength{\belowrulesep}{2pt}
    \begin{tabular}{crlcccccc|lcc}
    \toprule
        \multicolumn{9}{c|}{Zero-Shot SBIR} & \multicolumn{3}{c}{Cross-category Zero-Shot FG-SBIR}\\
        \cmidrule(lr){1-12} 
          \multicolumn{3}{c}{\multirow{2}{*}{Methods}} 
         & \multicolumn{2}{c}{Sketchy} 
         & \multicolumn{2}{c}{TU-Berlin} 
         & \multicolumn{2}{c|}{QuickDraw}
         & \multicolumn{1}{c}{\multirow{2}{*}{Methods}}
         & \multicolumn{2}{c}{Sketchy} \\
        \cmidrule(l){4-9}\cmidrule(lr){11-12} 
         & & & mAP@200 & P@200 & mAP@all & P@100 & mAP@all & P@200 & & Top-1 & Top-5 \\
        \cmidrule(lr){1-12}
        \multirow{8}{*}{\rotatebox[origin=l]{90}{ZS-SOTA}}
        & ECCV '18 & ZS-CAAE~\cite{yelamarthi2018zero}   & 0.156 & 0.260 & 0.005 & 0.003 & -- & --           \\%
        & ECCV '18 & ZS-CVAE~\cite{yelamarthi2018zero}   & 0.225 & 0.333 & 0.005 & 0.001 & 0.003 & 0.003           &  Cross-GRAD \cite{shankar2018generalizing} & 13.4 & 34.90  \\
        & CVPR '19 & ZS-CCGAN~\cite{dutta2019semantically}     & -- & -- & 0.297 & 0.426 & -- & --               \\
        & CVPR '19 & ZS-GRL~\cite{dey2019doodle}               & 0.369 & 0.370 & 0.110 & 0.121 & 0.075 & 0.068          & CC-DG \cite{pang2019generalising} & 22.6 & 49.00  \\ 
        \cmidrule(r){10-12} 
        & ICCV'19 & ZS-SAKE~\cite{liu2019semantic}            & 0.497 & 0.598 & 0.475 & 0.599 & -- & --           \\
        & AAAI '20 & ZS-GCN~\cite{zhang2020zero}               & 0.568 & 0.487 & 0.110 & 0.121 & -- & --                & B-FG-FT & 1.23 & 4.56  \\
        & NeurIPS '20 & ZS-IIAE~\cite{hwang2020variational}            & 0.373 & 0.485 & 0.412 & 0.503 & -- & --     \\
        & TPAMI '21 & ZS-TCN~\cite{wang2021transferable}            & 0.516 & 0.608 & 0.495 & 0.616 & 0.140 & 0.298     & B-FG-Lin & 15.75 & 39.63  \\
        & AAAI '22 & ZS-TVT~\cite{tian2022tvt}            & 0.531 & 0.618 & 0.484 & 0.662 & 0.149 & 0.293       \\
        & ACM MM '22 & ZS-PSKD[ViT]~\cite{wang2022prototype}            & 0.560 & 0.645 & 0.502 & 0.662 & 0.150 & 0.298  & B-FG-Cond\ & 25.98 & 54.38 \\
        & CVPR '22 & ZS-Sketch3T~\cite{sain2022sketch3t}    & 0.579 & 0.648 & 0.507 & 0.671 & -- & -- \\
        \cmidrule(l){1-9} 
        \multirow{6}{*}{\rotatebox[origin=l]{90}{B-CLIP}}
        & & B-FT                & 0.102 & 0.166 & 0.003 & 0.001 & 0.001 & 0.001   & B-FG-IP & 26.69 & 56.08  \\
        & & B-Lin               & 0.422 & 0.512 & 0.398 & 0.557 & 0.082 & 0.098  \\
        & & B-Cond              & 0.618 & 0.675 & 0.562 & 0.648 & 0.159 & 0.312   & B-FG-MM & 27.16 & 59.46 \\
        & & B-IP                & 0.691 & 0.711 & 0.628 & 0.702 & 0.182 & 0.361   \\
        & & B-MM                & 0.685 & 0.691 & 0.604 & 0.678 & 0.171 & 0.347  & B-FG-Deep & 27.62 & 61.56   \\
        & & B-Deep              & 0.702 & 0.718 & 0.637 & 0.718 & 0.188 & 0.375  \\
        \cmidrule(lr){1-12} 
        \rowcolor{Gray}
        & & \bf Ours & \bf 0.723 & \bf 0.725 & \bf 0.651 & \bf 0.732 & \bf 0.202 & \bf 0.388  & \bf Ours & \bf 28.68 & \bf 62.34   \\
        \bottomrule
    \end{tabular}
    \label{tab:main_table}
    \vspace{-0.4cm}
\end{table*}
}

\vspace{-0.1cm}
\subsection{Competitors}
\vspace{-0.15cm}
First we compare against \textbf{State-of-the-arts} for ZS-SBIR and FG-ZS-SBIR. For ZS-SBIR \textbf{(\textit{ZS}-SOTA)}, while \textit{ZS-CVAE} \cite{yelamarthi2018zero} and \textit{ZS-CAAE} \cite{yelamarthi2018zero} employs sketch-to-image translation, \textit{ZS-CCGAN} \cite{dutta2019semantically} and \textit{ZS-GRL} \cite{dey2019doodle} both use word2vec\cite{mikolov2013efficient} embeddings for semantic transfer, with adversarial learning and gradient-reversal layer respectively. 
Apart from using knowledge-distillation (KD) (\textit{ZS-SAKE}) \cite{liu2019semantic}, or learning a correlation matrix via prototype-based selective KD (\textit{ZS-PSKD} \cite{liu2019semantic}), complex designs like graph convolution network (\textit{ZS-GCN}), coupled sketch/photo encoder (\textit{ZS-TCN} \cite{wang2021transferable}) with shared conv layers but independent batchnorm layer, or complicated three-way ViT \cite{dosovitskiy2021image} architecture (\textit{ZS-TVT} \cite{tian2022tvt}) for visual/semantic transfer have been used. While \textit{ZS-IIAE} \cite{hwang2020variational} enforces cross-domain disentanglement, \textit{ZS-Sketch3T}\cite{sain2022sketch3t} uses a test-time training paradigm to minimise the train-test distribution gap. For FG-ZS-SBIR, we compare against \textit{CrossGrad} \cite{shankar2018generalizing} that leverages hard triplets with a category/domain classifier using word2vec embedded class-labels, and \textit{CC-DG} \cite{pang2019generalising} that models a universal manifold of prototypical visual sketch traits towards generalising to unseen categories. We report their results directly from their papers.

Next we design a few baselines (\textbf{B}) for adapting \textit{CLIP} to ZS-SBIR and ZS-FG-SBIR paradigms. 
For all of them, the prompt design remains same for both paradigms, but every baseline of ZS-SBIR (\textbf{B-}) is extended to FG-ZS-SBIR (\textbf{B-FG-}) across multiple categories using \textit{hard-triplets} and a CLIP text-encoder based classification loss.
\textit{B-FT} and \textit{B-FG-FT} fine-tune a pre-trained ViT-B/16 CLIP-Image Encoder \cite{clip}, for ZS-SBIR with a low learning rate of 1e-6. Similarly, \textit{B-Lin} and \textit{B-FG-Lin} use a linear probe \cite{clip} to train an additional feature embedding layer on top of pre-trained CLIP features to adapt to ZS-SBIR and FG-ZS-SBIR respectively, keeping image-extractor backbone frozen. Following \cite{zhou2022conditional}, \textit{B-Cond} and \textit{B-FG-Cond}, learns to generate a sketch-conditioned (for every sketch) prompt via a lightweight network (ResNet-18), when after concatenation with image-patch features are fed to the CLIP's image encoder. \textit{B-IP} and \textit{B-FG-IP} learn two independent \textit{shallow} prompts, for sketch and photo, that are injected in the first ViT \cite{dosovitskiy2021image} layer following \cite{zhou2022learning} for ZS-SBIR and FG-ZS-SBIR respectively.  Instead of independent sketch and photo prompts \textit{B-MM} and \textit{B-FG-MM} adapts \cite{maple2022} to learn a multi-modal prompt for both sketch and photo to ensure mutual synergy and discourage learning independent uni-modal solutions. More specifically, we learn a \textit{single} photo prompt from which the sketch prompt is obtained as a photo-to-sketch projection, via a linear layer. \textit{B-Deep} and \textit{B-FG-Deep} employ \textit{deep} prompting \cite{jia2022visual} learning $N=9$ prompts for sketch and photo, injected into the first $N$ layers of ViT \cite{dosovitskiy2021image} backbone. The last three types for both paradigms importantly differ from our method in keeping their LayerNorm frozen, which \textit{we} fine-tune for better accuracy. 

\vspace{-0.15cm}
\subsection{Performance Analysis:} 
\vspace{-0.2cm}
\keypoint{ZS-SBIR:} 
While state-of-the-arts offer reasonable performance (\Cref{tab:main_table}) thanks to their respective strategies of semantic transfer via word2vec (\textit{ZS-GRL}), adaptation via test-time training (\textit{ZS-Sketch3T}) or improvised transformer-based (\textit{ZS-TVT}), distillation-based (\textit{ZS-PSKD}) and other setups, our method armed with open-vocab generalisable potential of CLIP, surpasses them in all three datasets.
Although naively adapting large foundation models like CLIP \cite{clip} (e.g., \textit{B-FT}) understandably collapses, a successful adaptation outperforms existing SOTAs by $\approx24.8\%$ (avg). This motivates CLIP \cite{DenseCLIP} as the default sketch/photo encoders for future sketch research. While linear probing in \textit{B-Lin} secures higher results than SOTAs, it is surpassed by prompt-based learning, thus providing insights on a better adaptation choice. 
The marginal difference in performance between a simple adaptation of CLIP in \textit{B-IP} and more its complicated versions in \textit{B-Deep} \cite{jia2022visual}, \textit{B-MM}, \textit{B-Cond}, motivates the use of a simple \textit{shallow} prompt without the added ``bells and whistles" for ZS-SBIR. Finally, high accuracy on ZS-SBIR, establishes CLIP as a robust choice for sketch-based systems and thus motivates consequent focus on the more challenging task of cross-category FG-ZS-SBIR.

\keypoint{Cross-category FG-ZS-SBIR:} 
Our CLIP adapted paradigm easily surpasses the two existing SOTAs on cross-category FG-ZS-SBIR (\Cref{tab:main_table} right). While CLIP models \cite{DenseCLIP} shows impressive performance at category-level ZS-SBIR, there is reasonable scope for improvement in FG-ZS-SBIR that additionally requires structural matching\cite{pang2020solving}. Relatively higher improvement of \textit{B-FG-MM} over \textit{B-FG-IP} \cite{zhou2022learning} than its category-level counterpart (\textit{B-MM} over \textit{B-IP}), suggests the efficacy of multi-modal prompts over independent sketch/photo prompts, in FG-SBIR. This supports the prior observation \cite{yu2016sketch} that sharing encoder parameters is more suited to FG-SBIR \cite{sain2020cross} whereas separate sketch/photo weights work best at category-level. Lastly, learning a conditional prompt \cite{zhou2022conditional} in \textit{B-FG-Cond} offers marginal gain due to its sensitivity to training strategies \cite{maple2022, zhou2022conditional}.

\begin{figure*}
\centering
    \includegraphics[width=0.245\linewidth]{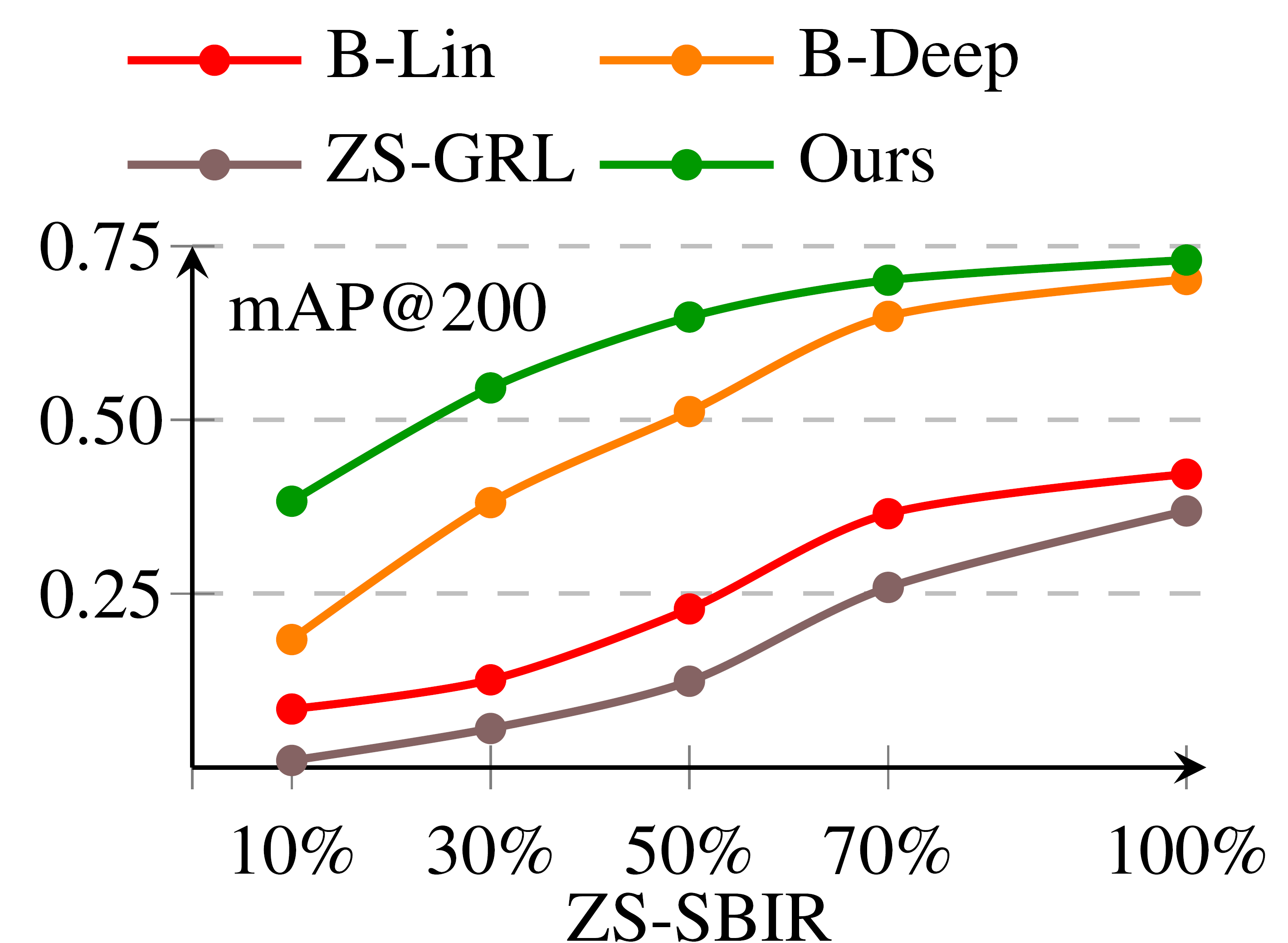}
    \includegraphics[width=0.245\linewidth]{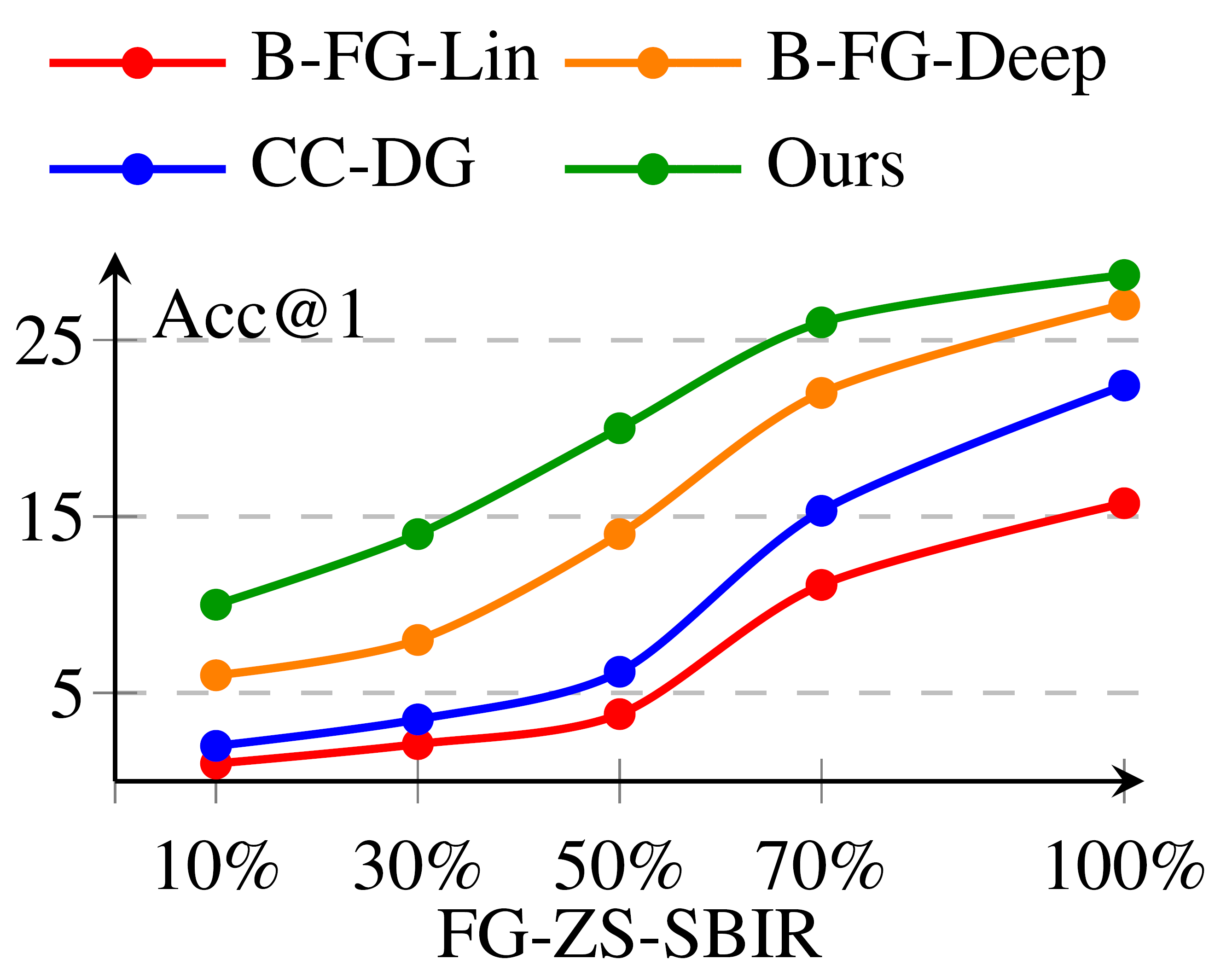}
    \includegraphics[width=0.245\linewidth]{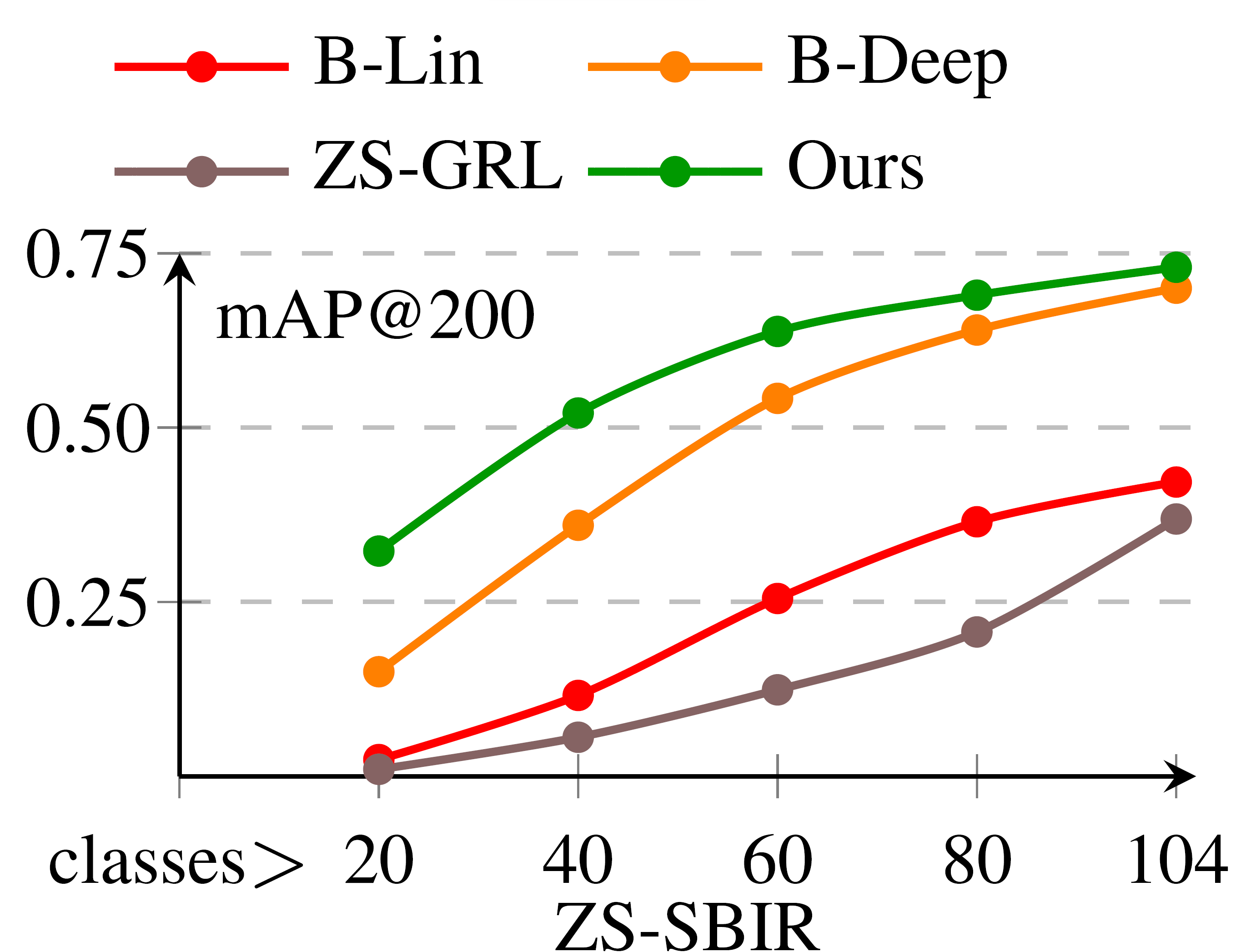}
    \includegraphics[width=0.245\linewidth]{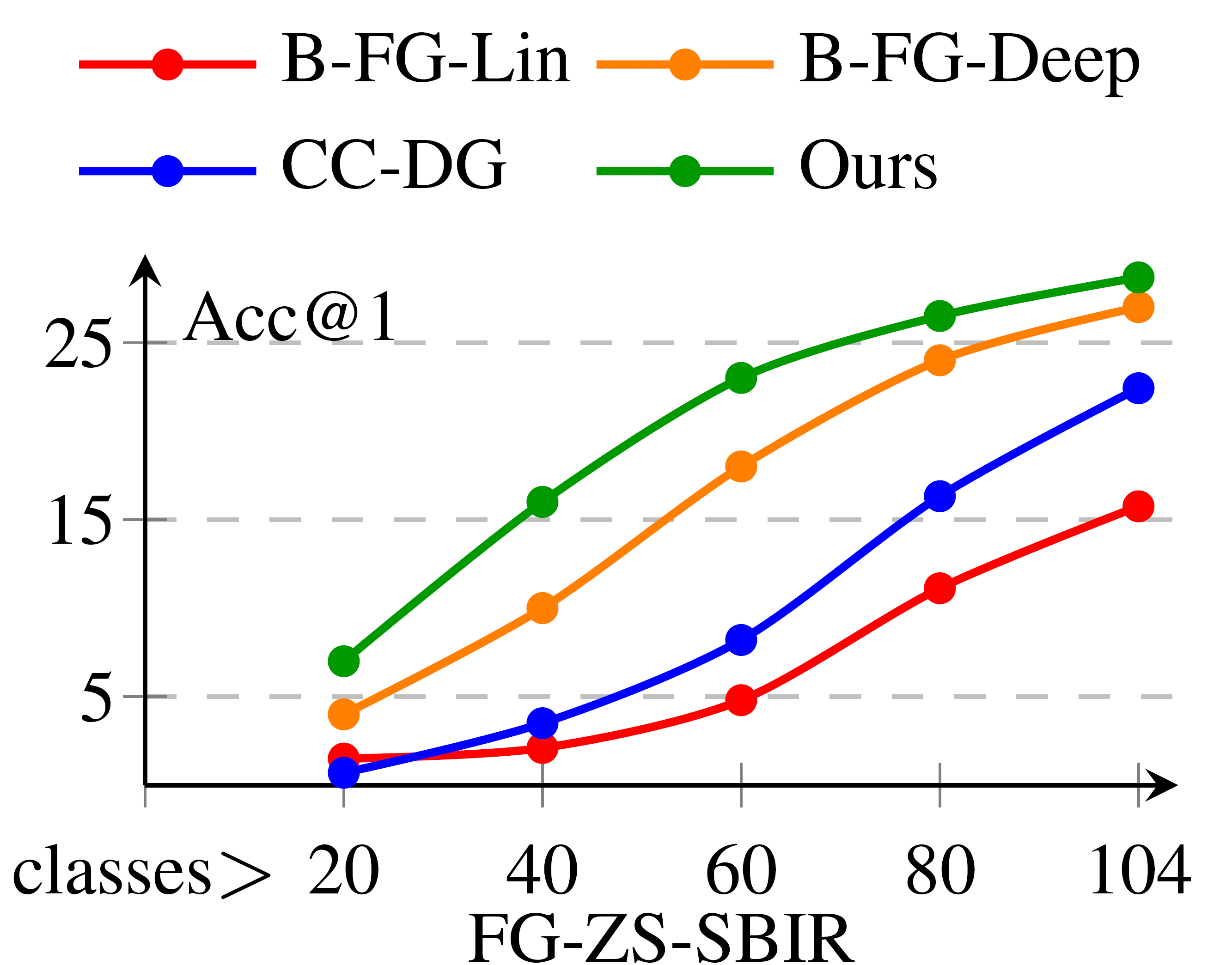}
    \vspace{-0.35cm}
	\caption{Plots showing extent of generalisation by varying training data-size (left two) and number of seen categories (right two).}
    \cut{x-axis we have the number of unseen categories -- increasing. For y-axis, we have the cross-category FG-SBIR performance of our method with others. We see with increasing unseen categories, CLIP based models gives a lower drop signifying a higher degree of generalisation to unseen and open-vocab setups.}
    \label{fig: generalisation}
    \vspace{-0.65cm}
\end{figure*}

\keypoint{Extent of Generalisation:} A strong motivation for using CLIP is its out-of-distribution performance \cite{DenseCLIP} that promises to enable large-scale real-world deployment of sketch applications. Scrutinising this generalisation potential further, we experiment in two paradigms: \textit{(i)} we vary training data per-class as 10\%, 30\%, 50\%, 70\% and 100\%, and (ii) we vary number of seen classes as 20, 40, 60, 80 and 104 from Sketchy \cite{sangkloy2016sketchy} respectively. \cref{fig: generalisation} shows our ZS-SBIR and FG-ZS-SBIR performance to remain relatively stable at variable training-data-size (left) as well as across variable number of seen (training) categories (right), compared to prior arts and baselines, justifying the zero-shot potential of CLIP for both ZS-SBIR and FG-ZS-SBIR tasks.

\vspace{-0.15cm}
\subsection{Ablation Study}
\label{sec:ablation}
\vspace{-0.2cm}


\keypoint{Justifying design components:} We evaluate our models, dropping one component at a time (\Cref{tab:ablation}) for both ZS-SBIR and FG-ZS-SBIR. While not fine-tuning LayerNorm (\textit{w/o LayerNorm}) lowers performance on both paradigms slightly, removing $\mathcal{L}^\mathcal{I}_\text{cls}$ (\textit{w/o f-Divergene}) severely affects FG-ZS-SBIR as it loses it class-discrimination ability. Removing $\mathcal{L}_\text{PS}$ (\textit{w/o Patch-Shuffling}) and $\mathcal{L}_\delta$ (\textit{w/o f-Divergence}) lowers FG-ZS-SBIR accuracy by 3.15\% and 3.75\%, owing to loss of sketch-photo structural correspondence and non-uniform relative sketch-photo distances across categories, thus verifying contribution of every design choice. Furthermore, using $2\times2$ patches instead of $3\times3$ for FG-ZS-SBIR provides a 1.2\% gain in Acc.@1 (Sketchy), thus being optimal, as larger grid means more white patches for sketch, leading to confused embeddings.

\vspace{-0.2cm}
\begin{table}[!htbp]
\setlength{\tabcolsep}{5pt}
\renewcommand{\arraystretch}{0.85}
\footnotesize
\centering
\caption{Ablation Study on Sketchy}
\vspace{-0.25cm}
\label{tab:quant_sBIR}
\begin{tabular}{lcccc}
\toprule
{\multirow{2}{*}{Methods}} & \multicolumn{2}{c}{ZS-SBIR} & \multicolumn{2}{c}{FG-ZS-SBIR} \\
\cmidrule(lr){2-3}\cmidrule(lr){4-5}
& mAP@all & P@200 & Top-1 & Top-5  \\  
\cmidrule(lr){1-1}\cmidrule(lr){2-3}\cmidrule(lr){4-5}
w/o LayerNorm      & 0.698 & 0.701 &  27.18 & 59.55 \\  
w/o Classification ($\mathcal{L}^\mathcal{I}_\text{cls}$)   & 0.703 & 0.710 &  10.69 & 16.32  \\ [0.05cm] 
w/o Patch-Shuffling ($\mathcal{L}_\text{PS}$)  & -- & -- &  25.18  & 53.07 \\ [0.05cm] 
w/o f-Divergence ($\mathcal{L}_\delta$)      & -- & -- &  24.93 & 53.72  \\
\cmidrule(lr){1-5}
\rowcolor{Gray}
\bf Ours  & \bf 0.723 & \bf 0.725 & \bf 28.68 & \bf 62.34\\
\bottomrule
\end{tabular}
\vspace{-0.1cm}
\label{tab:ablation}
\end{table}

\keypoint{CLIP text encoder v/s Word2vec:} Unlike earlier works \cite{dey2019doodle} that obtained semantics of a category via reconstruction from word2vec \cite{mikolov2013efficient} embeddings for that category, our method replaces it with feature embedding from CLIP's text encoder \cite{clip}. Word2vec \cite{mikolov2013efficient} is trained using text-\textit{only} data, whereas CLIP's text encoder trains on large-scale ($400M$) image-text \textit{pairs}. Using word2vec embeddings in our proposed method instead, drops performance by $4.57\%/0.172$ (Acc@1/mAP@200) for ZS-SBIR/FG-ZS-SBIR on Sketchy \cite{sangkloy2016sketchy}, justifying our choice of CLIP's text encoder, capturing visual-semantic association (text-photo pairs) instead of text-only information (word2vec).

\keypoint{Should we \textit{learn} text prompts?:} While handcrafted prompt templates like `\texttt{a photo of a [category]}' works well for class-discrimination training (Eqn. \ref{eq: f-divergence}), we wish to explore if there is any need to \textit{learn} text prompts like our visual prompts. Consequently, following \cite{zhou2022conditional} we take $N$ learnable prompts, matching the word embeddings of handcrafted prompt dimensionally for the $i^\text{th}$ class as:  $\eta^t = \{\eta_1,\eta_2,\cdots, \eta_N,c_i\}$ with word embedding $c_i$ denoting the class name \cite{zhou2022conditional}. 
Training accordingly, drops performance by $2.36\%/0.078$ (Acc.@1/mAP@200) in Sketchy for ZS-SBIR/ZS-SBIR. This is probably because unlike learned prompts, \textit{handcrafted} prompts being rooted in \textit{natural language} vocabulary are inclined to have a higher generalisation ability in our case than learned text-prompts \cite{mikolov2013efficient}.

\keypoint{Varying number of Prompts: } To ensure that our visual prompt ($\mathbb{R}^{K \times d_{p}}$) has sufficient information capacity \cite{tishbyIB}, we experiment with $K$ = $\{1, 2, 3, 4\}$. Accuracy improves from $26.15\%/0.675$ at $K$ = 1 to $28.68\%/0.723$ at $K$ = 3, but saturates to $28.26\%/0.718$ (Acc@1/mAP@200) at $K$ = 4 on Sketchy, proving optimal $K$ as 3. 
We conjecture that a high capacity prompt might lead to over-fitting \cite{zhou2022learning} of CLIP thus resulting in a lower zero-shot performance.

\keypoint{Comparing text-based image retrieval:}
To explore how sketch fares against keywords as a query in a zero-shot retrieval paradigm, we compare keyword-based retrieval against ZS-SBIR on Sketchy(ext), and against FG-ZS-SBIR on Song \etal's \cite{song2017fine} dataset having fine-grained sketch-photo-text triplets for fine-grained retrieval. While keyword based retrieval employed via off-the-shelf CLIP, remains competitive ($0.523/0.612$ mAP/P@200) against our ZS-SBIR framework ($0.723/0.725$), it lags behind substantially ($4.6\%$ Acc@1) from our ZS-FG-SBIR method ($18.68\%$ Acc@1), proving the well-accepted superiority of sketch in modelling fine-grained details over text \cite{yu2016sketch}.

\vspace{-0.3cm}
\section{Conclusion}
\label{sec:conclusion}
\vspace{-0.2cm}
In  this work we leverage CLIP's open-vocab generalisation potential via an intuitive prompt-based design to enhance zero-shot performance of SBIR -- both category-level and fine-grained, where our method surpasses all prior state-of-the-arts significantly. Towards improving fine-grained ZS-SBIR, we put forth two novel strategies of making relative sketch-photo feature distances across categories uniform and learning structural sketch-photo correspondences via a patch-shuffling technique. Last but not least, we hope to have informed the sketch community on the potential of synergizing foundation models like CLIP and sketch-related tasks going forward.


{\small
\bibliographystyle{ieee_fullname}
\bibliography{main}
}

\cleardoublepage
\onecolumn{%
\begin{center}
\title{\Large \textbf{Supplementary material for \\CLIP for All Things Zero-Shot Sketch-Based Image Retrieval, \\ Fine-Grained or Not}}
\vspace{0.2cm}
\author{
\\Aneeshan Sain\textsuperscript{1,2}  \hspace{.2cm} 
Ayan Kumar Bhunia\textsuperscript{1} \hspace{.2cm}  
Pinaki Nath Chowdhury\textsuperscript{1,2} \hspace{.2cm}
Subhadeep Koley\textsuperscript{1,2} \hspace{.2cm}\\
Tao Xiang\textsuperscript{1,2}\hspace{.4cm}  
Yi-Zhe Song\textsuperscript{1,2} \\
\textsuperscript{1}SketchX, CVSSP, University of Surrey, United Kingdom.  \\
\textsuperscript{2}iFlyTek-Surrey Joint Research Centre on Artificial Intelligence.\\
\tt\small{\{a.sain, a.bhunia, p.chowdhury, s.koley, t.xiang, y.song\}@surrey.ac.uk}
}  
\end{center}
}

\renewcommand\thesection{\Alph{section}}
\setcounter{section}{0}

\section{Prompt Design for ZS-SBIR and ZS-FG-SBIR}Here we show in detail the way visual prompts are incorporated into the Image Encoder of CLIP \cite{clip}.

\begin{figure}[!hbt]
    \centering
    \includegraphics[width=\linewidth]{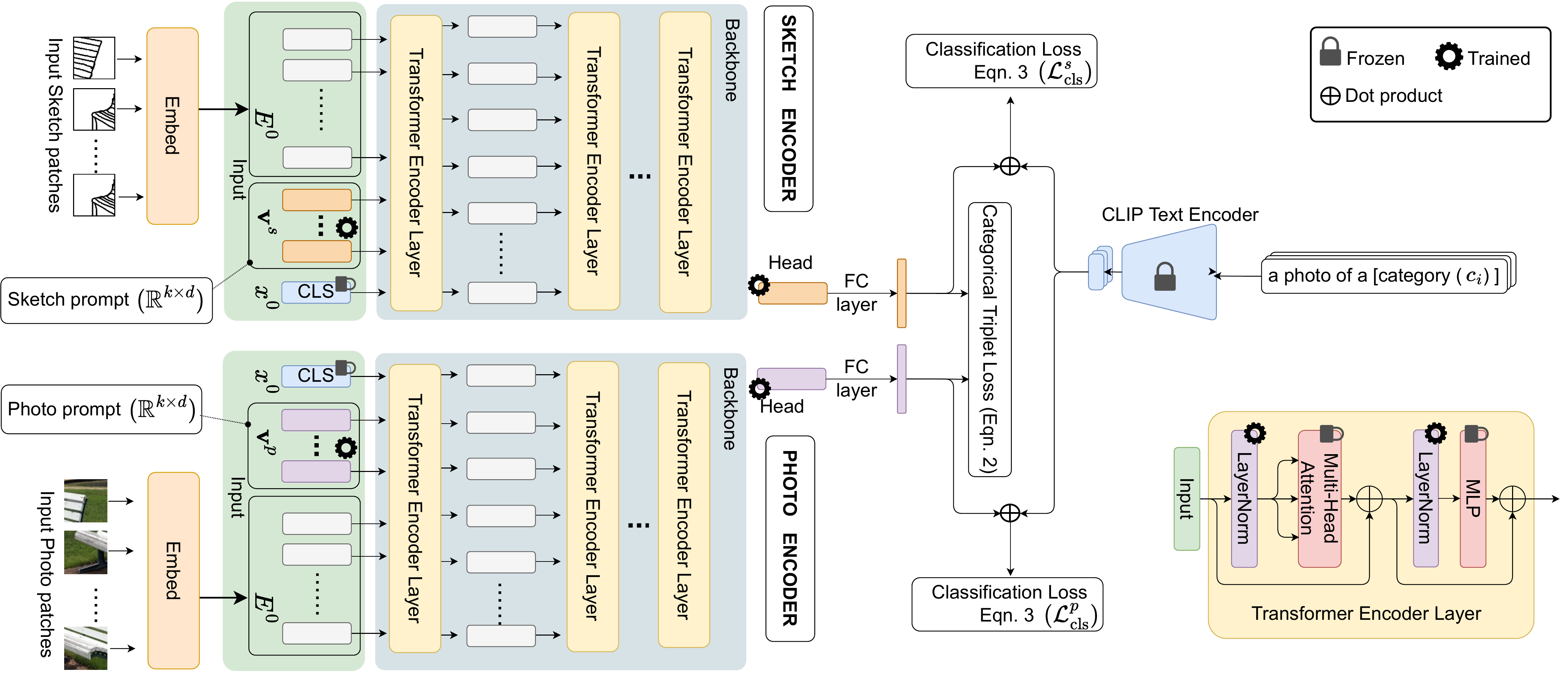}
    \caption{Prompt Design for ZS-SBIR along with training objectives.}
    \label{fig:ZSSBIR-prompt}
\end{figure}

For \textbf{ZS-SBIR}, we have two separate CLIP-image-encoders for photo and sketch branch with sketch and photo prompts incorporated into the respective encoders. During training the entire CLIP model is kept frozen except the LayerNorm of transformer layers and the prompts themselves, as shown in \cref{fig:ZSSBIR-prompt}.

\begin{figure}[!hbt]
    \centering
    \includegraphics[width=\linewidth]{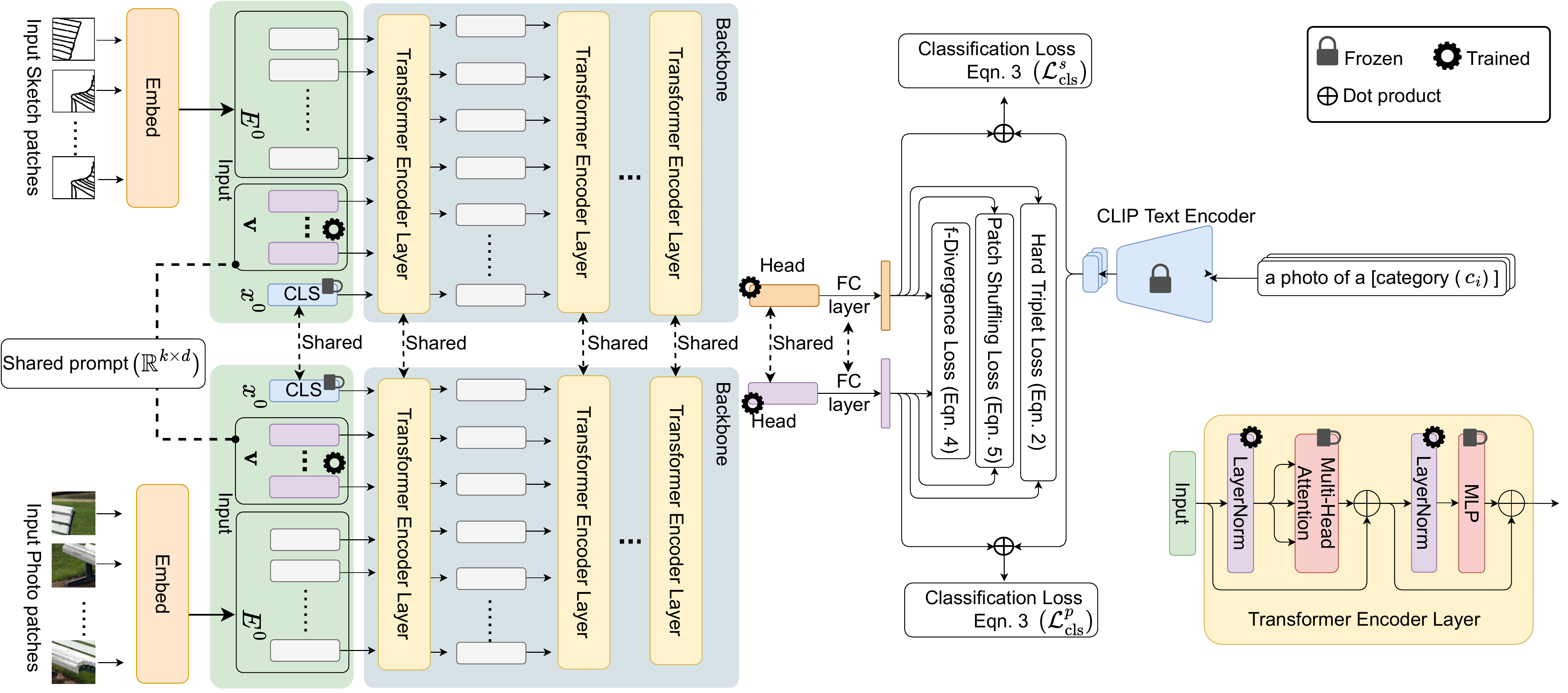}
    \vspace{-0.5cm}
    \caption{Prompt Design for FG-ZS-SBIR along with training objectives.}
    \label{fig:FGZSSBIR-prompt}
\end{figure}

For \textbf{FG-ZS-SBIR}, we use one CLIP-image-encoder with \textit{one common prompt} shared for both photo and sketch branches incorporated into the CLIP-image encoder. Similar to ZS-SBIR design, during training the entire CLIP model is kept frozen except the LayerNorm of transformer layers and the prompt itself, as shown in \cref{fig:FGZSSBIR-prompt}. Apart from shared image-encoders the main difference from ZS-SBIR is in considering hard-triplets within each category instead of category-level triplets as in ZS-SBIR. Furthermore, we have two more additional losses apart from the ones used for ZS-SBIR, aimed at (i) making the relative sketch-photo distances across categories uniform via f-divergence, and (ii) learning the structural correspondences between a sketch-photo pair via Patch-shuffling loss.

\section{Datasets}For evaluation on \textbf{ZS-SBIR} we have used three datasets:

(i) \textbf{Sketchy (extended)} \cite{liu2017deep} --  Sketchy \cite{sangkloy2016sketchy} contains 75,471 sketches over 125 categories having 100 images per category, with atleast 5 associated hand-drawn sketches per photo \cite{yelamarthi2018zero}. It was extended \cite{liu2017deep} further with extra 60,502 images from ImageNet \cite{russakovsky2015imagenet} (Sketchy-ext), which we use here. Following \cite{yelamarthi2018zero} for zero-shot setup we split it as 104 classes for training and 21 for testing, ensuring that \textit{test}-set images do not overlap with 1000 classes of ImageNet~\cite{russakovsky2015imagenet}.

(ii) \textbf{TUBerlin \cite{eitz2012humans}} -- contains 250 categories, with 80 free-hand sketches in each, which was  extended with a total of 204,489 images by \cite{zhang2016sketchnet}. We split it following \cite{dey2019doodle} as 30 classes for testing and 220 for training. 

(iii) \textbf{QuickDraw Extended} -- The full-version contains over 50 million sketches across 345 categories, drawn by users across the internet under 20 seconds per sketch. Augmenting the sketches with images from \textit{Flickr}, a subset of QuickDraw with 110 categories having 330,000 sketches and 204,000 photos was introduced for ZS-SBIR in \cite{dey2019doodle}. We follow their split of 80 classes for training and 30 for testing to ensure no overlap of test-set photos from ImageNet \cite{russakovsky2015imagenet}.

\begin{figure}[!hbt]
    \centering
    \includegraphics[width=\linewidth]{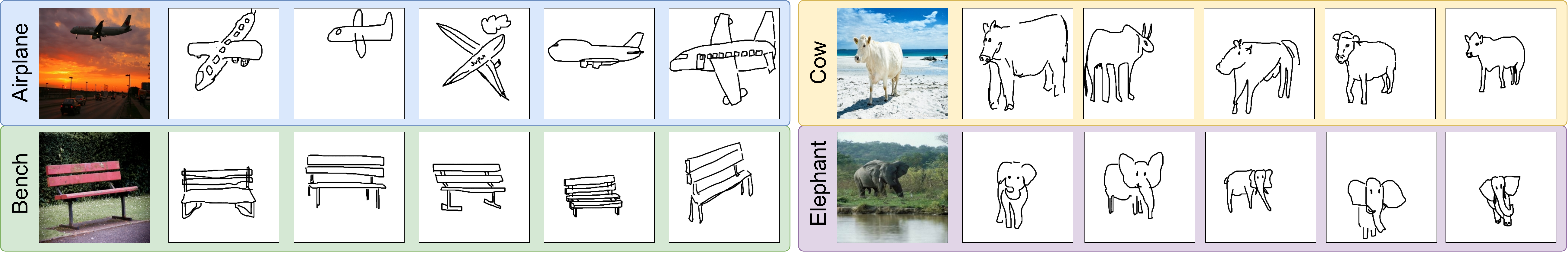}
    \caption{Some examples of Fine-grained associations across categories from Sketchy \cite{sangkloy2016sketchy}.}
    \label{fig:datasetFG}
\end{figure}

\vspace{+0.2cm}
For evaluation on \textbf{FG-ZS-SBIR} we require fine-grained (one-to-one matching) sketch-photo association \cite{yu2016sketch} across categories for evaluation. Accordingly we resort to Sketchy \cite{sangkloy2016sketchy} which has atleast atleast 5 associated hand-drawn sketches associated to every photo \cite{yelamarthi2018zero}. We use the same zero-shot categorical split of 104 training and 21 testing classes \cite{yelamarthi2018zero}. A few examples of sketch-photo association with multiple sketches per photo is illustrated in \cref{fig:datasetFG}.

\section{More on f-Divergence} To use a single (global) margin parameter $\mu$ that works for all categories, we impose a regulariser that aims to make the sketch-photo relative distance, defined as $\delta(s, p^{+}, p^{-})$, uniform across categories. We achieve this by computing the distribution of relative distances for all triplets $(s, p^{+}, p^{-})$ in category $c$ as $\mathcal{D}_{c} = \{ \delta(s_i, p^{+}_i, p^{-}_i) \}_{i=1}^{N_{s}} $, where the $c^{th}$ category has $N_{s}$ sketch-photo pairs. Next, towards making the relative distance uniform across all categories, we minimise the KL-divergence \cite{kl-divergence} between a distribution of relative distances.

However, KL-divergence \cite{kl-divergence} only computes the information distance between two distributions -- the length of the shortest program to describe a second distribution given the first. Comparing multiple $(\geq 2)$ distributions however is comparatively less studied. The multi-distribution generalisation of information distance, aka., the $f$-\textit{divergence} is defined by a convex function $f: [0, \infty) \rightarrow \mathbb{R}$. Despite its generalisation capability, $f$-divergence for multiple distribution setup is under-explored in computer vision applications. In this paper, we thus adopt a rather simplistic definition of $f$-divergence by Sgarro \cite{sgarro1981} -- the average divergence, which is defined as,

\begin{equation}
    \frac{1}{N_s(N_s-1)} \sum_{i=1}^{N_s} \sum_{j=1}^{N_s} \texttt{KL} (\mathcal{D}_{i}, \mathcal{D}_{j})
\end{equation}

\section{Some Qualitative Results on Sketchy} 
\begin{figure}[!hbt]
    \centering
    \includegraphics[width=\linewidth]{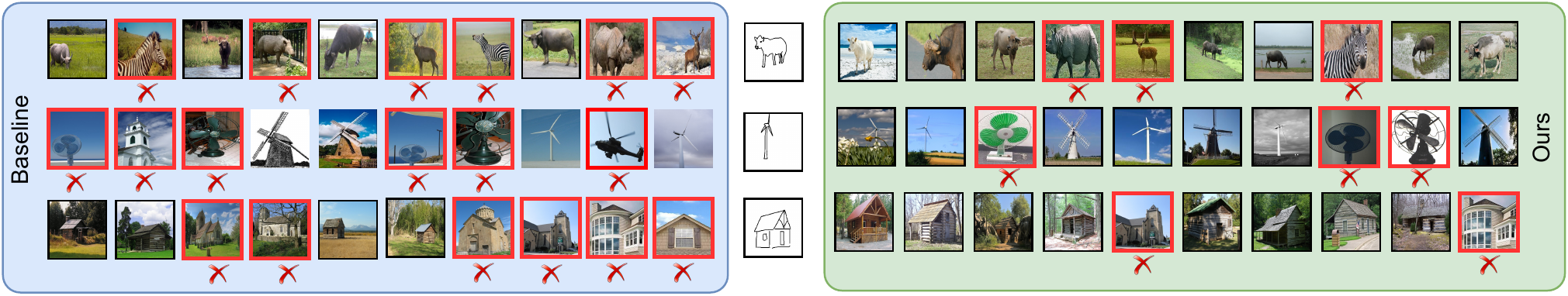}
    \caption{Qualitative results of ZS-SBIR on Sketchy \cite{liu2017deep} by a {baseline} (\textcolor{myBlue}{blue}) method vs {Ours} (\textcolor{deepGreen}{green}). }
    \label{fig:qual-SBIR}
    \vspace{-0.4cm}
\end{figure}

\begin{figure}[!hbt]
    \centering
    \includegraphics[width=\linewidth]{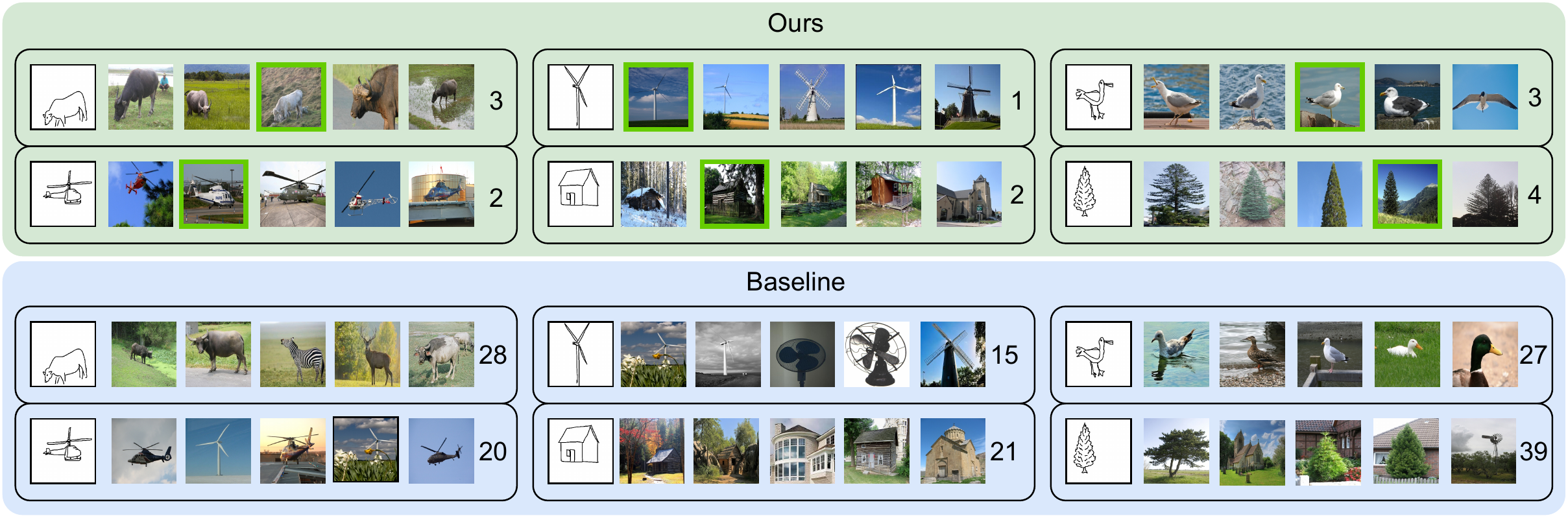}    
    \hspace*{-0.22cm}\includegraphics[width=0.7\linewidth]{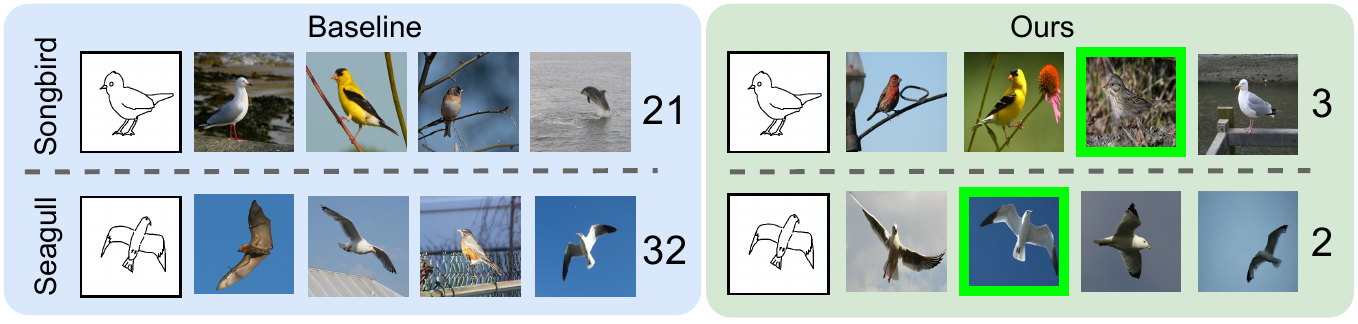}
    \caption{Qualitative results of FG-ZS-SBIR on Sketchy \cite{sangkloy2016sketchy} by a {baseline} (\textcolor{myBlue}{blue}) method vs {Ours} (\textcolor{deepGreen}{green}). The images are arranged in increasing order of the ranks beside their corresponding sketch-query, i.e the left-most image was retrieved at rank-1 for every category. The true-match for every query, if appearing in top-5 is marked in a green frame. Numbers denote the rank at which that true-match is retrieved for every corresponding sketch-query.}
    \label{fig:qual-FGSBIR}
    \vspace{-0.4cm}
\end{figure}

Figures show qualitative results on Sketchy (ext) \cite{liu2017deep} for ZS-SBIR (\cref{fig:qual-SBIR}) and on Sketchy \cite{sangkloy2016sketchy} for FG-ZS-SBIR (\cref{fig:qual-FGSBIR}), of baseline methods vs. ours. Baselines are constructed following \cite{dey2019doodle} and \cite{yu2016sketch} for ZS-SBIR and FG-ZS-SBIR respectively.

\section{Limitations} We observed two plausible limitations of our method which we keep for addressal in a future work. \textit{(i)} The assumption that CLIP covers almost all classes during training, might fail in certain niche cases. \textit{(ii)} Being trained on internet-scale data (400M image-text pairs), thorough zero-shot evaluation on an unseen class is challenging. However both limitations are universal to all CLIP-based applications.

\end{document}